\documentclass[10pt,twocolumn,letterpaper]{article}

\usepackage{cvpr}
\usepackage{times}
\usepackage{epsfig}
\usepackage{graphicx}
\usepackage{amsmath}
\usepackage{amssymb}
\usepackage{algorithm}
\usepackage{multirow}
\usepackage{mathtools,xparse}
\usepackage{subcaption}
\usepackage{siunitx}
\usepackage{mathtools}
\usepackage{stmaryrd}
\usepackage[noend]{algpseudocode}
\usepackage{placeins}

\newcommand*\rot{\rotatebox{90}}
\newcommand*\trans{\text{T}}

\newcommand*\Points{\mathcal{P}}
\newcommand*\PointsDesc{\mathcal{P}^*}

\usepackage[pagebackref=true,breaklinks=true,letterpaper=true,colorlinks,bookmarks=false]{hyperref}

\cvprfinalcopy % *** Uncomment this line for the final submission

\def\cvprPaperID{361} % *** Enter the iccv Paper ID here

\ifcvprfinal\fi
\begin{document}

%%%%%%%%% TITLE
\title{Multi-Class Model Fitting by Energy Minimization and Mode-Seeking}
% Multi-X: Energy Minimization and Mode-Seeking for Multi-Class Model Recovery

\author{Daniel Barath\\
Machine Perception Research Laboratory\\
MTA SZTAKI, Budapest, Hungary\\
{\tt\small barath.daniel@sztaki.mta.hu}
\and
Jiri Matas\\
Centre for Machine Perception, Department of Cybernetics \\
Czech Technical University, 
Prague, Czech Republic\\
{\tt\small matas@cmp.felk.cvut.cz}
}

\maketitle

\newtheorem{mydef}{Definition}

%%%%%%%%% ABSTRACT
\begin{abstract}
	We propose a general formulation, called Multi-X, for multi-class multi-instance model fitting -- the problem of interpreting the input data as a mixture of noisy observations originating from multiple instances of multiple classes. We extend the commonly used $\alpha$-expansion-based technique with a new move in the label space. The move replaces a set of labels with the corresponding density mode in the model parameter domain, thus achieving fast and robust optimization.   
Key optimization parameters like the bandwidth of the mode seeking are set automatically within the algorithm. Considering that a group of outliers may form spatially coherent structures in the data, we propose a cross-validation-based technique removing statistically insignificant instances. 
Multi-X outperforms significantly the state-of-the-art on publicly available datasets for diverse problems: multiple plane and rigid motion detection; motion segmentation; simultaneous plane and cylinder fitting; circle and line fitting.

%Multi-X outperforms significantly the state-of-the-art on the standard AdelaideRMF (multiple plane segmentation, multiple rigid motion detection) and Hopkins datasets (motion segmentation) and in experiments on 3D LIDAR data (simultaneous plane and cylinder fitting) and on 2D edge interpretation (circle and line fitting). Multi-X runs in time approximately linear in the number of data points at around 0.1 second per 100 points, an order of magnitude faster than available implementations of commonly used methods. 
\end{abstract}

%%%%%%%%% BODY TEXT
\section{Introduction}
\begin{figure}
      \centering
      \fbox{\begin{minipage}{0.40\columnwidth}\includegraphics[width = 1.0\columnwidth]{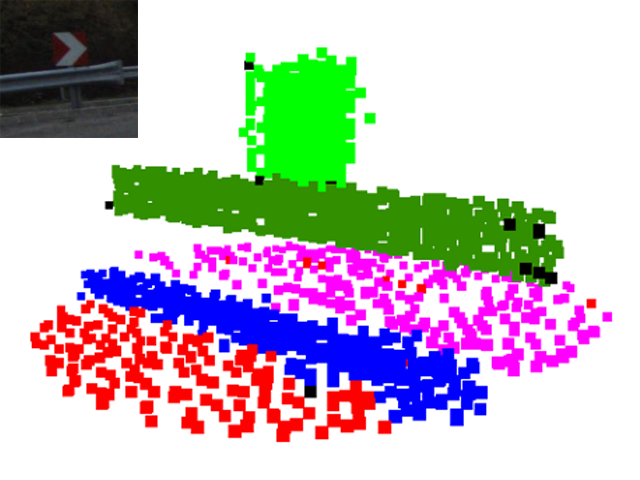}\end{minipage}}
      \fbox{\begin{minipage}{0.40\columnwidth}\includegraphics[width = 1.0\columnwidth]{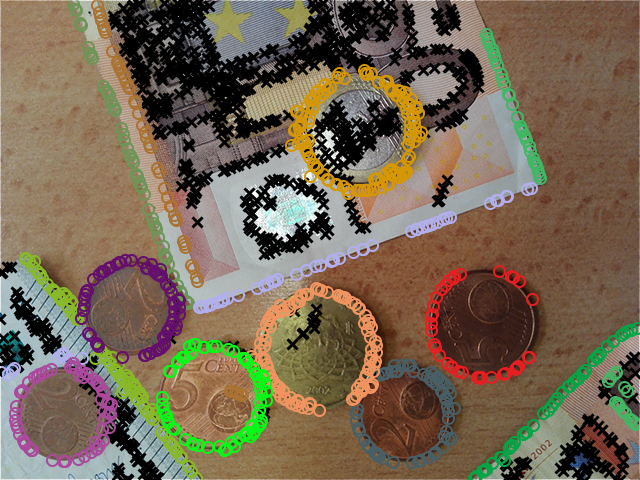}\end{minipage}}\\
      \fbox{\begin{minipage}{0.40\columnwidth}\includegraphics[width = 1.0\columnwidth]{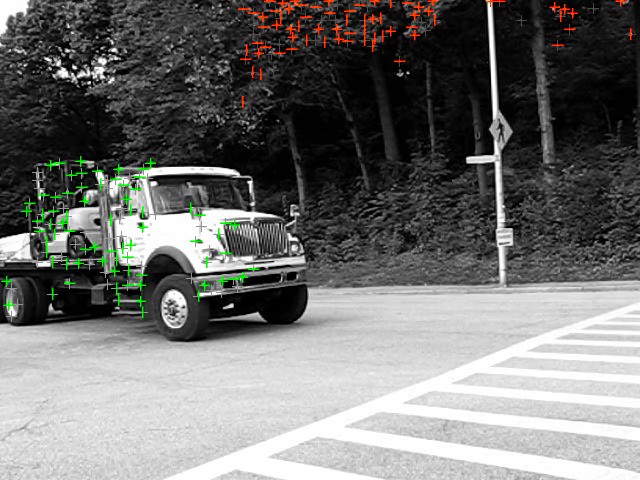}\end{minipage}}
      \fbox{\begin{minipage}{0.40\columnwidth}\includegraphics[width = 1.0\columnwidth]{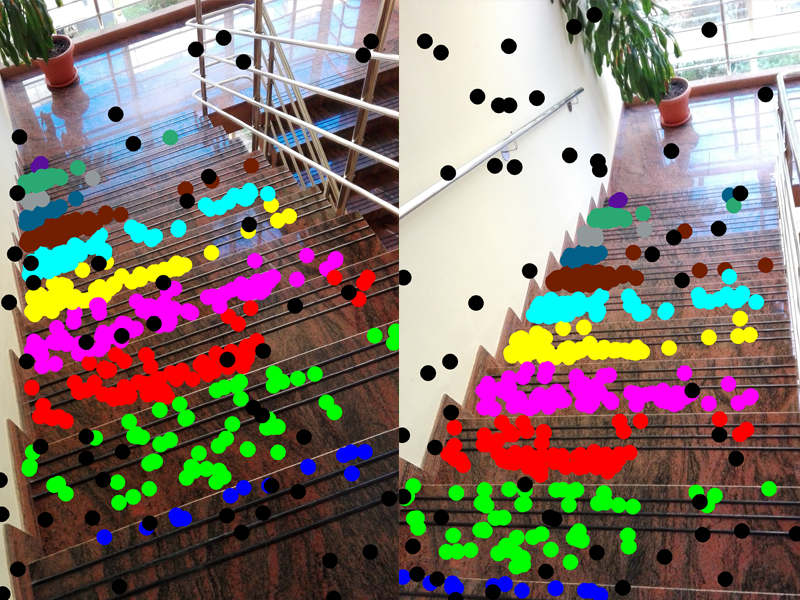}\end{minipage}}
      \caption{ Multi-class multi-instance fitting examples. Results on simultaneous plane and cylinder (top left), line and circle fitting (top right), motion (bottom left) and plane segmentation (bottom right).}
      \label{fig:example_problems}
\end{figure}

In multi-class fitting, the input data is interpreted as a mixture of noisy observations originating from multiple instances of multiple model classes, e.g.\ $k$ lines and $l$ circles in 2D edge maps, $k$ planes and $l$ cylinders in 3D data, multiple homographies or fundamental matrices from correspondences from a non-rigid scene (see Fig.~\ref{fig:example_problems}). Robustness is achieved by considering assignment to an  outlier class. 

Multi-model fitting has been studied since the early sixties, the Hough-transform~\cite{vc1962method,illingworth1988survey} being the first popular method for extracting multiple instances of a single class~\cite{guil1997lower,matas2000robust,rosin1993ellipse,xu1990new}. A widely used approach for finding a single instance is RANSAC~\cite{fischler1981random} which alternates two steps: the generation of instance hypotheses and their validation. However, extending RANSAC to the multi-instance case has had limited success.
Sequential RANSAC detects instance one after another in a greedy manner, removing their inliers~\cite{vincent2001detecting,kanazawa2004detection}. In this approach, data points are assigned to the first instance, typically the one with the largest support for which they cannot be deemed outliers, rather than to the best instance.  MultiRANSAC~\cite{zuliani2005multiransac} forms compound hypothesis about $n$ instances. Besides requiring the number $n$ of the instances to be known a priori, the approach increases the size of the minimum sample and thus the number of hypotheses that have to be validated.  

Most recent approaches~\cite{isack2012energy,magri2014t,magri2015robust,magri2016multiple,toldo2008robust} focus on the single class case: finding multiple instances of the same model class. A popular group of methods~\cite{delong2012minimizing,isack2012energy,pham2014interacting,pham2014random,barath2016multi} adopts a two step process: initialization by RANSAC-like instance generation followed by a point-to-instance assignment optimization by \textit{energy minimization} using graph labeling techniques~\cite{boykov2004experimental}.
%
%Recently, an approach was proposed combining Mean-Shift and energy minimization to solve the two-view homography fitting problem in rigid scenes~\cite{barath2016multi}.
%
Another group of methods uses \textit{preference analysis}, introduced by RHA~\cite{zhang2007nonparametric}, which is based on  the distribution of residuals of individual data points with respect to the instances~\cite{magri2014t,magri2015robust,toldo2008robust}. 

The \textit{multiple instance multiple class case} considers fitting of instances that are not necessarily of the same class. This generalization has received much less attention than the single-class case. To our knowledge, the last significant contribution is that of Stricker and Leonardis~\cite{stricker1995exsel} who search for multiple parametric models simultaneously by minimizing description length using Tabu search. 

The proposed Multi-X method finds multiple instances of multiple model classes drawing on progress in energy minimization extended with a new move in the label space: replacement of a set of labels with the corresponding density mode in the model parameter domain. Mode seeking significantly reduces the label space, thus speeding up the energy minimization, and it overcomes the problem of multiple instances with similar parameters, a weakness of state-of-the-art single-class approaches. The assignment of data to instances of different model classes is handled by the introduction of class-specific distance functions. Multi-X can also be seen as an extension or generalization of the Hough transform: (i) it finds modes of the parameter space density without creating an accumulator and locating local maxima there, which is prohibitive in high dimensional spaces, (ii) it handles multiple classes -- running Hough transform for each model type in parallel or sequentially cannot easily handle competition for data points, and (iii) the ability to model spatial coherence of inliers and to consider higher-order geometric priors is added.

Most recent papers~\cite{magri2014t,magri2016multiple,wang2015mode} report results tuned for each test case separately. The results are impressive, but input-specific tuning, i.e.\ semi-automatic operation with multiple passes, severely restricts possible applications.
We propose an \textit{adaptive parameter setting} strategy within the algorithm, allowing the user to run Multi-X as a black box on a range of problems with no need to set any parameters. 
Considering that outliers may form structures in the input, as a post-processing step, a cross-validation-based technique removes insignificant instances. 

The contributions of the paper are: 
(i) A general formulation is proposed for multi-class multi-instance model fitting which, to the best of our knowledge, has not been investigated before. 
(ii) The commonly used energy minimizing technique, introduced by PEARL~\cite{isack2012energy}, is extended with a new move in the label space: replacing a set of labels with the corresponding density mode in the model parameter domain. Benefiting from this move, the minimization is speeded up, terminates with lower energy and the estimated model parameters are more accurate.  
(iii) The proposed pipeline combines state-of-the-art techniques, such as energy-minimization, median-based mode-seeking, cross-validation, to achieve results superior to the recent multi-model fitting algorithms both in terms of accuracy and processing time. Proposing automatic setting for the key optimization parameters, the method is applicable to various real world problems. 

\section{Multi-Class Formulation}

Before presenting the general definition, let us consider a few examples of multi-instance fitting: to find a pair of \textit{line instances} $h_1, h_2 \in \mathcal{H}_l$ interpreting a set of 2D points $\Points \subseteq \mathbb{R}^2$. Line class $\mathcal{H}_l$ is the space of lines $\mathcal{H}_l = \{(\theta_l, \phi_l, \tau_l), \theta_l = [\alpha \quad c]^\trans \}$ equipped with a distance function $\phi_l(\theta_l, p) = |\cos(\alpha) x + \sin(\alpha) y + c|$ ($p~=~[x \quad y]^\trans~\in~\Points$) and a function $\tau_l(p_1, ..., p_{m_l}) = \theta_l$ for estimating $\theta_l$ from $m_l \in \mathbb{N}$ data points. Another simple example is the fitting $n$ \textit{circle instances} $h_1, h_2, \cdots, h_n \in \mathcal{H}_c$ to the same data. The circle class $\mathcal{H}_c = \{(\theta_c, \phi_c, \tau_c), \theta_c = [c_x \quad c_y \quad r]^\trans\}$ is the space of circles, $\phi_c(\theta_c, p) = |r - \sqrt{(c_x - x)^2 + (c_y - y)^2}|$ is a distance function and $\tau_c(p_1, ..., p_{m_c}) = \theta_c$ is an estimator.
\textit{Multi-line fitting} is the problem of finding multiple line instances $\{h_1, h_2, ...\} \subseteq \mathcal{H}_l$, while the \textit{multi-class} case is extracting a subset $\mathcal{H} \subseteq \mathcal{H}_\forall$, where $\mathcal{H}_\forall = \mathcal{H}_l \cup \mathcal{H}_c \cup \mathcal{H}_. \cup \cdots$. The set $\mathcal{H}_\forall$ is the space of all classes, e.g.\ line and circle. The formulation includes the outlier class $\mathcal{H}_o = \{(\theta_o, \phi_o, \tau_o), \theta_o = \emptyset \}$ where each instance has constant but possibly different distance to all points $\phi_o(\theta_o, p) = k$, $k~\in~\mathbb{R}^+$ and $\tau_o(p_1, ..., p_{m_o}) = \emptyset$. Note that considering multiple outlier classes allows interpretation of outliers askk originating from different sources.  

\begin{mydef}[Multi-Class Model]
	The multi-class model is a space $\mathcal{H}_\forall = \bigcup \mathcal{H}_i$, where $\mathcal{H}_i = \{ (\theta_i, \phi_i, \tau_i) \; | \; d_i \in \mathbb{N}, \theta_i \in \mathbb{R}^{d_i}, \phi_i \in \Points \times \mathbb{R}^{d_i} \rightarrow \mathbb{R}, \tau_i : \PointsDesc \to \mathbb{R}^{d_i} \}$ is a single class, $\Points$ is the set of data points, $d_i$ is the dimension of parameter vector $\theta_i$, $\phi_i$ is the distance function and $\tau_i$ is the estimator of the $i$th class. 
\end{mydef}

The \textit{objective of multi-instance multi-class model fitting} is to determine a set of instances $\mathcal{H} \subseteq \mathcal{H}_\forall$ and labeling $L \in \Points \rightarrow \mathcal{H}$ assigning each point $p \in \Points$ to an instance $h \in \mathcal{H}$ minimizing energy $E$. 
We adopt energy 
\begin{equation}
	\label{eq:old_overall_energy}
	E(L) = E_d(L) + w_g E_g(L) + w_c E_c(L)
\end{equation} 
to measure the quality of the fitting, where $w_g$ and $w_c$ are weights balancing the different terms described bellow, and $E_d$, $E_c$ and $E_g$ are the data, complexity terms, and the one considering geometric priors, e.g.\ spatial coherence or perpendicularity, respectively.

\noindent
$\textbf{Data term}$ $E_d: (\Points \rightarrow \mathcal{H}) \rightarrow \mathbb{R}$  is defined in most energy minimization approaches as 
\begin{equation}
	\label{eq:data_term}
	E_d(L) = \sum_{p \in \Points} \phi_{L(p)}(\theta_{L(p)}, p),
\end{equation}
penalizing inaccuracies induced by the point-to-instance assignment, where $\phi_{L(p)}$ is the distance function of $h_{L(p)}$.  

\noindent
\textbf{Geometric prior term} $E_g$ considers spatial coherence of the data points, adopted from~\cite{isack2012energy}, and possibly higher order geometric terms~\cite{pham2014interacting}, e.g.\ perpendicularity of instances. The term favoring spatial coherence, i.e.\ close points more likely belong to the same instance, is defined as  
\begin{equation}
	\label{eq:spatial_term}
	E_g(L) : (\Points \rightarrow \mathcal{H}) \rightarrow \mathbb{R} = \sum_{(p,q) \in N} w_{pq} \llbracket L(p) \not= L(q) \rrbracket,
\end{equation}
where $N$ are the edges of a predefined neighborhood-graph, the Iverson bracket $\llbracket.\rrbracket$ equals to one if the condition inside holds and zero otherwise, and $w_{pq}$ is a pairwise weighting term. In this paper, $w_{pq}$ equals to one. For problems, where it is required to consider higher-order geometric terms, e.g.\ to find three perpendicular planes, $E_g$ can be replaced with the energy term proposed in~\cite{pham2014interacting}.

\noindent
\textbf{A regularization of the number of instances} is proposed by Delong et al.~\cite{delong2012fast} as a label count penalty
%
%\begin{equation*}
$
	E_c(L) : (\Points \rightarrow \mathcal{H}) \rightarrow \mathbb{R} =  |L(\Points)|,  
$
%\end{equation*}
%
where $L(\Points)$ is the set of distinct labels of labeling function $L$. 
To handle multi-class models which might have different costs on the basis of the model class, we thus propose the following definition:  
\begin{mydef}[Weighted Multi-Class Model]
	The weighted multi-class model is a space $\widehat{\mathcal{H}}_\forall = \bigcup \widehat{\mathcal{H}}_i$, where $\widehat{\mathcal{H}}_i = \{ (\theta_i, \phi_i, \tau_i, \psi_i) \; | \; d_i \in \mathbb{N}, \theta_i \in \mathbb{R}^{d_i}, \phi_i \in \Points \times \mathbb{R}^{d_i} \rightarrow \mathbb{R}, \tau_i : \PointsDesc \to \mathbb{R}^{d_i},  \psi_i \in \mathbb{R} \}$ is a weighted class, $\Points$ is the set of data points, $d_i$ is the dimension of parameter vector $\theta_i$, $\phi_i$ is the distance function, $\tau_i$ is the estimator, and $\psi_i$ is the weight of the $i$th class. 
\end{mydef}
The term controlling the number of instances is
\begin{equation}
	\label{eq:complexity_term}
	\widehat{E}_c(L) = \sum_{l \in L(\Points)} \psi_{l},  
\end{equation}
instead of $E_c$, where $\psi_{l}$ is the weight of the weighted multi-class model referred by label $l$.  

\noindent
Combining terms Eqs.~\ref{eq:data_term},~\ref{eq:spatial_term},~\ref{eq:complexity_term} leads to \textbf{overall energy} $\widehat{E}(L) = E_d(L) + w_g E_g(L) + w_c \widehat{E}_c(L)$. 

\section{Replacing Label Sets}

For the optimization of the previously described energy, we build on and extend the PEARL algorithm~\cite{isack2012energy}. PEARL generates a set of initial instances applying a RANSAC-like randomized sampling technique, then alternates two steps until convergence: \vspace{0.6em}

\noindent
(1) Application of $\alpha$-expansion~\cite{boykov2001fast} to obtain labeling $L$ minimizing overall energy $\widehat{E}$ w.r.t.\ the current instance set. \vspace{0.6em}

\noindent
(2) Re-estimation of the parameter vector $\theta$ of each model instance in $\mathcal{H}$ w.r.t.\ labeling $L$.\vspace{0.6em}

\noindent
In the PEARL formulation, the only way for a label to be removed, i.e.\ for an instance to be discarded, is to assign it to no data points. Experiments show that (i) this removal process is often unable to delete instances having similar parameters, (ii) and makes the estimation sensitive to the choice of label cost $w_c$. We thus propose a new move in the label space: replacing a set of labels with the density mode in the model parameter domain. 

\begin{figure}[htbp]
	\centering
	\includegraphics[width = 0.45\columnwidth]{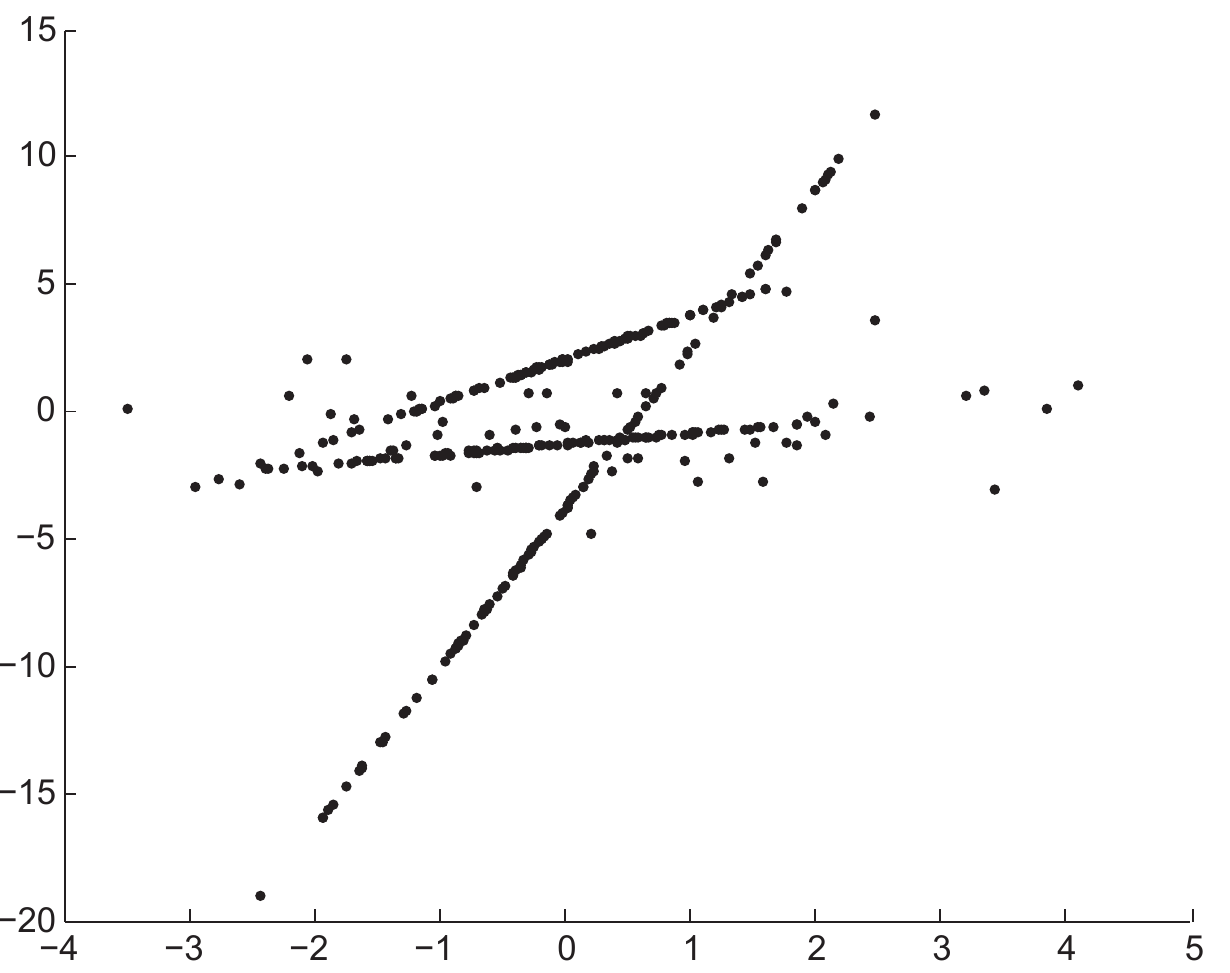}
    \includegraphics[width = 0.45\columnwidth]{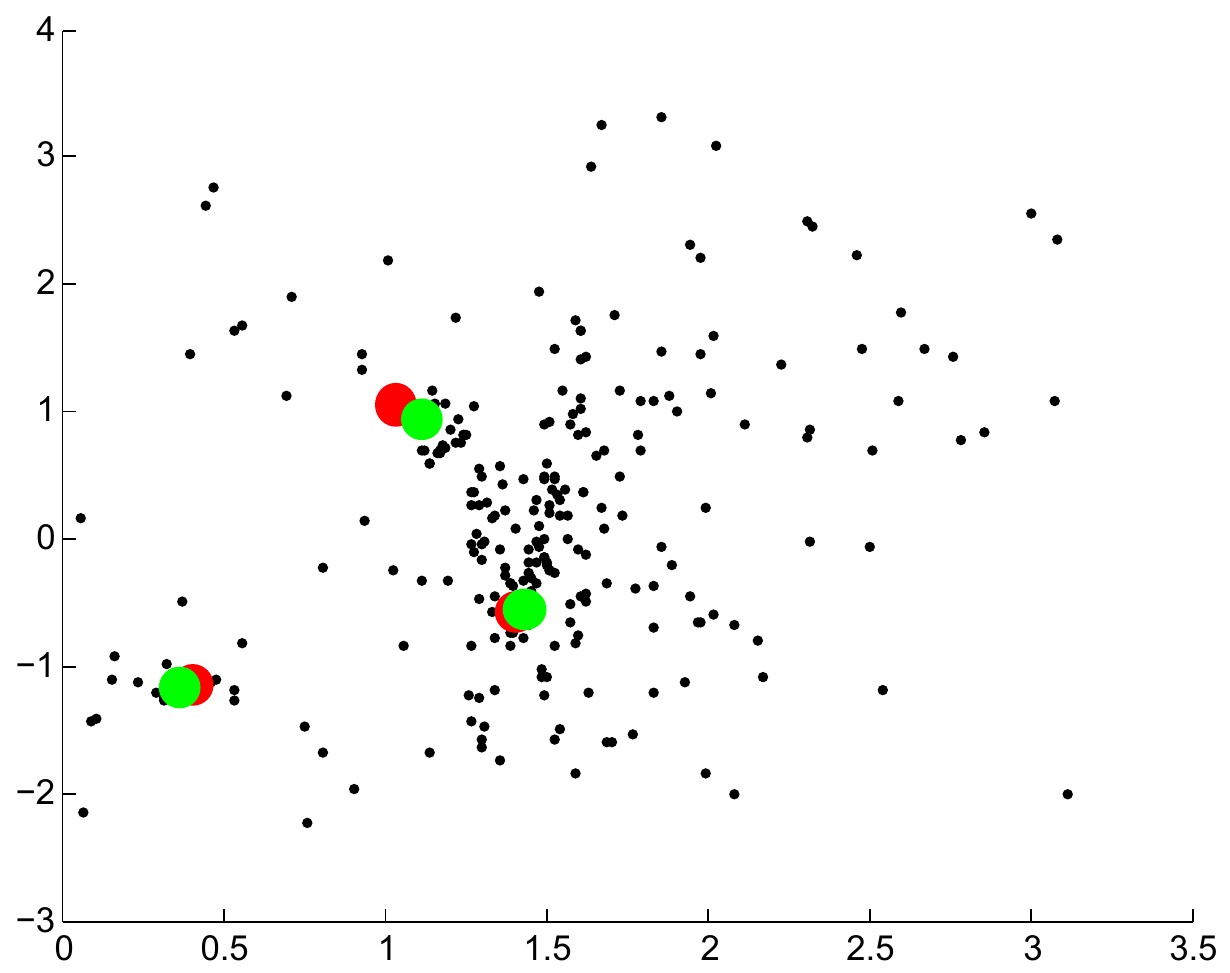}
	\caption{ (\textbf{Left}) Three lines each generating $100$ points with zero-mean Gaussian noise added, plus $50$ outliers. (\textbf{Right}) $1000$ line instances generated from random point pairs, the ground truth instance parameters (red dots) and the modes (green) provided by Mean-Shift shown in the model parameter domain: $\alpha$ angle -- vertical, offset -- horizontal axis. }
	\label{fig:mode_seeking}
\end{figure}

Multi-model fitting techniques based on energy-minimization usually generate a high number of instances $\mathcal{H} \subseteq \mathcal{H}_\forall$ randomly as a first step~\cite{isack2012energy,pham2014interacting} ($|\mathcal{H}| \gg |\mathcal{H}_{\text{real}}|$, where $\mathcal{H}_{\text{real}}$ is the ground truth instance set). Therefore, the presence of many similar instances is typical. 
We assume, and experimentally validate, that many points supporting the sought instances in $\mathcal{H}_{\text{real}}$ 
are often assigned in the initialization to a number of  instances in $\mathcal{H}$ with similar parameters. The cluster around the ground truth instances in the model parameter domain can be replaced with the modes of the density (see Fig.~\ref{fig:mode_seeking}). 

Given a mode-seeking function $\Theta: \mathcal{H}_\forall^* \rightarrow \mathcal{H}_\forall^*$, e.g.\ Mean-Shift~\cite{comaniciu2002mean}, which obtains the density modes of input instance set $\mathcal{H}_i$ in the $i$th iteration. The proposed move is as
\begin{equation}
	\mathcal{H}_{i+1} := \begin{cases*}
		\Theta(\mathcal{H}_{i}) & if $E(L_{\Theta(\mathcal{H}_{i})}) \leq E(L_{i})$,  \\
        \mathcal{H}_{i} & otherwise, 
	\end{cases*}
    \label{eq:mode_move}
\end{equation}
where $L_i$ is the labeling in the $i$th iteration and $L_{\Theta(\mathcal{H}_{i})}$ is the optimal labeling which minimizes the energy w.r.t.\ to instance set $\Theta(\mathcal{H}_{i})$. It can be easily seen, that Eq.~\ref{eq:mode_move} does not break the convergence since it replaces the instances, i.e.\ the labels, \textit{if and only if} the energy does not increase. Note that clusters with cardinality one -- modes supported by a single instance -- can be considered as outliers and removed. This step reduces the label space and speeds up the process.

\section{Multi-X}

The proposed approach, called Multi-X, combining PEARL, multi-class models and the proposed label replacement move, is summarized in Alg.~\ref{alg:multi_x}. Next, each step is described. 

%Therefore the proposed approach combining PEARL with a move in the a new step is added to the alternation of PEARL as a first one: .
%\noindent
%In Multi-X, these steps are alternated until convergence. Next, each step is described in depth (see Alg.~\ref{alg:multi_x}). 
\begin{algorithm}
\begin{algorithmic}[1]
	\Statex{\hspace{-1.0em}\textbf{Input:} $P$ -- data points}
    \Statex{\hspace{-1.0em}\textbf{Output:} $H^*$ -- model instances, $L^*$ -- labeling}
   	\Statex{}
    \State{$H_0$ := InstanceGeneration($P$); $i$ := $1$;}
   	\Repeat
    	\State{$H_{i}$ := ModeSeeking$(H_{i - 1})$;} \Comment by Median-Shift
    	\State{$L_{i}$ := Labeling($H_{i}$, $P$);}  \Comment by $\alpha$-expansion
    	%\State{$\gamma$ := AdaptiveOutlierThreshold($P$, $H_0$);}
       %\State{$L_{i}$ := OutlierRemoval($H_{i}$, $L_{i}$, $\gamma$);}
        \State{$H_{i}$ := ModelFitting($H_{i}$, $L_{i}$, $P$);} \Comment by Weiszfeld
      %  \State{$Convergence$ := CheckConvergence();}
        \State{$i$ := $i + 1$;}
    \Until{$!$Convergence($H_{i}$, $L_{i}$)}
    \State{$H^*$ := $H_{i-1}$, $L^*$ := $L_{i-1}$;}
    \State{$H^*$, $L^*$ := ModelValidation($H^*$, $L^*$)} \Comment Alg.~\ref{alg:weak_model_removal}
\end{algorithmic}
\caption{\bf Multi-X}
\label{alg:multi_x}
\end{algorithm}\vspace{0.6em}

\paragraph{1. Instance generation} step generates a set of initial instances before the alternating optimization is applied. Reflecting the assumption that the data points are spatially coherent, we use the guided sampling of NAPSAC~\cite{nasuto2002napsac}. This approach first selects a random point, then the remaining ones are chosen from the neighborhood of the selected point. The same neighborhood is used as for the spatial coherence term in the $\alpha$-expansion. Note that this step can easily be replaced by e.g.\ PROSAC~\cite{chum2005matching} for problems where the spatial coherence does not hold or favors degenerate estimates, e.g.\ in fundamental matrix estimation.

\paragraph{2. Mode-Seeking} is applied in the model parameter domain. Suppose that a set of instances $\mathcal{H}$ is given. Since the number of instances in the solution -- the modes in the parameter domain -- is unknown, a suitable choice for mode-seeking is the Mean-Shift algorithm~\cite{comaniciu2002mean} or one of its variants. In preliminary experiments, the most robust choice was the Median-Shift~\cite{shapira2009mode} using Weiszfeld-~\cite{weiszfeld1937point} or Tukey-medians~\cite{tukey1975mathematics}. There was no significant difference, but Tukey-median was slightly faster to compute.
In contrast to Mean-Shift, it does not generate new elements in the vector space since it always return an element of the input set. With the Tukey-medians as modes, it is more robust than Mean-Shift~\cite{shapira2009mode}. However, we replaced Locality Sensitive Hashing~\cite{datar2004locality} with Fast Approximated Nearest Neighbors~\cite{muja2009fast} to achieve higher speed. 

Reflecting the fact that a general \textit{instance-to-instance} distance is needed, we represent instances by point sets, e.g.\ a line by two points and a homography by four correspondences, and define the \textit{instance-to-instance} distance as the Hausdorff distance~\cite{rockafellar2009variational} of the point sets. Even though it yields slightly more parameters than the minimal representation, thus making Median-Shift a bit slower, it is always available as it is used to define spatial neighborhood of points.
Another motivation for representing by points is the fact that having a non-homogeneous representation, e.g.\ a line described by angle and offset, leads to anisotropic distance functions along the axes, thus complicating the distance calculation in the mode-seeking.

There are many point sets defining an instance and a canonical point set representation is needed. For lines, the nearest point to the origin is used and a point on the line at a fixed distance from it. For a homography $\mathbf{H}$, the four points are $\mathbf{H} [0, 0, 1]^\trans$, $\mathbf{H} [1, 0, 1]^\trans$, $\mathbf{H} [0, 1, 1]^\trans$, and $\mathbf{H} [1, 1, 1]^\trans$. The matching step is excluded from the Hausdorff distance, thus speeding up the distance calculation significantly.\footnote{Details on the choice of model representation are provided in the supplementary material.}

The application of Median-Shift $\Theta_{\text{med}}$ never increases the number of instances $|\mathcal{H}_i|$:
%\begin{equation*}	
$
	|\Theta_{\text{med}}(\mathcal{H}_{i})| \leq |\mathcal{H}_i|.
$
%\end{equation*}
The equality is achieved \textit{if and only if} the distance between every instance pair is greater than the bandwidth. 
Note that for each distinct model class, Median-Shift has to be applied separately. According to our experience, applying this label replacement move in the first iteration does not make the estimation less accurate but speeds it up significantly even if the energy slightly increases. 

%
%After this step, the outlier threshold is re-estimated as discussed in Section~\ref{sec:adaptive_outlier_threshold}.\vspace{0.6em}
%(TODO: check)

\paragraph{3. Labeling} assigns points to model instances obtained in the previous step. A suitable choice for such task is $\alpha$-expansion~\cite{boykov2001fast}, since it handles an arbitrary number of labels. Given $\mathcal{H}_i$ and an initial labeling $L_{i-1}$ in the $i$th iteration, labeling $L_{i}$ is estimated using $\alpha$-expansion minimizing energy $\widehat{E}$. Note that $L_{0}$ is determined by $\alpha$-expansion in the first step. The number of the model instances $|\mathcal{H}_i|$ is fixed during this step and the energy must decreases: 
%\begin{equation*}	
$
	\widehat{E}(L_i, \mathcal{H}_i) \leq \widehat{E}(L_{i-1}, \mathcal{H}_i).
$
%\end{equation*}
%
To reduce the sensitivity on the outlier threshold (as it was shown for the single-instance case in~\cite{lebeda2012fixing}), the distance function of each class is included into a Gaussian-kernel.

%\noindent
%\textbf{3. The Outlier Removal} part of the algorithm determines points fitting no instance. A common technique to remove outliers in energy minimization tasks is to add an outlier label -- several approaches are discussed in~\cite{isack2009spatially}. 
%We experienced that our separation of $\alpha$-expansion and outlier removal steps does not affect the accuracy (it is described in Section~\ref{section:convergence} in depth). However, it facilitates the adaptive parameter setting. All points are marked outlier for which $\phi_p(\theta_p, p) > \gamma_p$, where $\gamma_p$ is a threshold corresponding to the instance for which point $p \in \Points$ is assigned to.\footnote{From the algorithmic point of view, this approach needs no additional processing time -- the cost of point assignment is stored during the optimization -- unlike dealing with an outlier label.} \vspace{0.6em}

\paragraph{4. Model Fitting} re-estimates the instance parameters w.r.t.\ the assigned points. The obtained instance set $\mathcal{H}_i$ is re-fitted using the labeling provided by $\alpha$-expansion. The number of the model instances $|\mathcal{H}_i|$ is constant. $L_2$ fitting is an appropriate choice, since combined with the labeling step, it can be considered as truncated $L_2$ norm. 
%
% We use the Weiszfeld-algorithm~\cite{weiszfeld1937point}, an iteratively re-weighted least squares approach, to achieve $L_1$ model fitting. Note that if the processing time is crucial, $L_2$ fitting is an appropriate choice, however, $L_1$ minimization is more robust. 

The overall energy $\widehat{E}$ can only decrease or stay constant during this step since it consists of three terms: (1) $E_d$ -- the sum of the assignment costs minimized, (2) $E_g$ -- a function of the labeling $L_i$, fixed in this step and (3) $\widehat{E}_c$ -- which depends on $|H_i|$ so $\widehat{E}_c$ remains the same. Thus
\begin{equation}	
	\label{eq:e_change}
	\widehat{E}(L_i, \mathcal{H}_{i+1}) \leq \widehat{E}(L_{i}, \mathcal{H}_{i}).
\end{equation}

\paragraph{5. Model Validation} 

considers that a group of outliers may form spatially coherent structures in the data. We propose a post-processing step to remove statistically insignificant models using cross-validation. The algorithm, summarized in Alg.~\ref{alg:weak_model_removal}, selects $ a minimal subsett$ times from the inlier points $I$. In each iteration, an instance is estimated from the selected points and its distance to each point is computed. The original instance is considered stable if the mean of the distances is lower than threshold $\gamma$. Note that $\gamma$ is the outlier threshold used in the previous sections.
\begin{algorithm}
\begin{algorithmic}[1]
	\Statex{\hspace{-1.0em}\textbf{Input:} $I$ -- inlier points, $t$ -- trial number, }
	\Statex{$\gamma$ -- outlier threshold} \Comment default $t = 100$
    \Statex{\hspace{-1.0em}\textbf{Output:} $R \in \{\text{true}, \text{false}\}$ -- response}
   	\Statex{}
    \State{$\widehat{D}$ := $0$}
   	\For{$i := 1 \textrm{ to } t$}
    	\State{MSS := SelectMinimalSubset($I$)} 
        \State{$H$ := ModelEstimation(MSS)}
    	\State{$\widehat{D}$ := $\widehat{D} +$ MeanDistanceFromPoints($H$, $I$) $/ t$}
        %\State{$D$ := MeanDistanceFromPoints($H$, $I$)}
    	%\State{$\widehat{D}$ := $\widehat{D} + D / t$}
    \EndFor
    \State{$R$ := $\widehat{D} < \gamma$}
\end{algorithmic}
\caption{\bf Model Validation.}
\label{alg:weak_model_removal}
\end{algorithm}
\paragraph{Automatic parameter setting} is crucial for Multi-X to be applicable to various real world tasks without requiring the user to set most of the parameters manually. 
To avoid manual bandwidth selection for \textbf{mode-seeking}, we adopted the automatic procedure proposed in~\cite{georgescu2003mean} which sets the bandwidth $\epsilon_i$ of the $i$th instance to the distance of the instance and its $k$th neighbor. Thus each instance has its own bandwidth set automatically on the basis of the input. 

\textbf{Label cost} $w_c$ is set automatically using the approach proposed in~\cite{pham2014interacting} as follows: $w_c = m \log(|\mathcal{P}|) / h_{\text{max}}$, where $m$ is the size of the minimal sample to estimate the current model, $|\mathcal{P}|$ is the point number and $h_{\text{max}}$ is the maximum expected number of instances in the data. 
Note that this cost is not required to be high since mode-seeking successfully suppresses instances having similar parameters. The objective of introducing a label cost is to remove model instances with weak supports. In practice, this means that the choice of $h_{\text{max}}$ is not restrictive.

Experiments show that the choice of the \textbf{number of initial instances} does not affect the outcome of Multi-X significantly. In our experiments, the number of instances generated was twice the number of the input points. 

Spatial coherence weight $w_g$ value $0.3$ performed well in  the experiments. The common problem-specific outlier thresholds which led to the most accurate results was: homographies ($2.4$ pixels), fundamental matrices ($2.0$ pixels), lines and circles ($2.0$ pixels), rigid motions ($2.5$), planes and cylinders ($10$ cm).

\section{Experimental Results}

First we compare Multi-X with PEARL~\cite{isack2012energy} combined with the label cost of Delong et al.~\cite{delong2012fast}. Then the performance of Multi-X applied to the following Computer Vision problems is reported: line and circle fitting, 3D plane and cylinder fitting to LIDAR point clouds, multiple homography fitting, two-view and video motion segmentation.
\vspace{0.2em}

\paragraph{Comparison of PEARL and Multi-X.}

In a test designed to show the effect of the proposed label move, model validation was not applied and both methods used the same algorithmic components described in the previous section. A synthetic environment consisting of three 2D lines, each sampled at $100$ random locations, was created. Then 200 outliers, i.e.\ random points, were added. 

\begin{figure*}[htbp]
	\centering
	\begin{subfigure}[t]{0.485\textwidth}
		\centering
    	\includegraphics[width = 0.45\columnwidth]{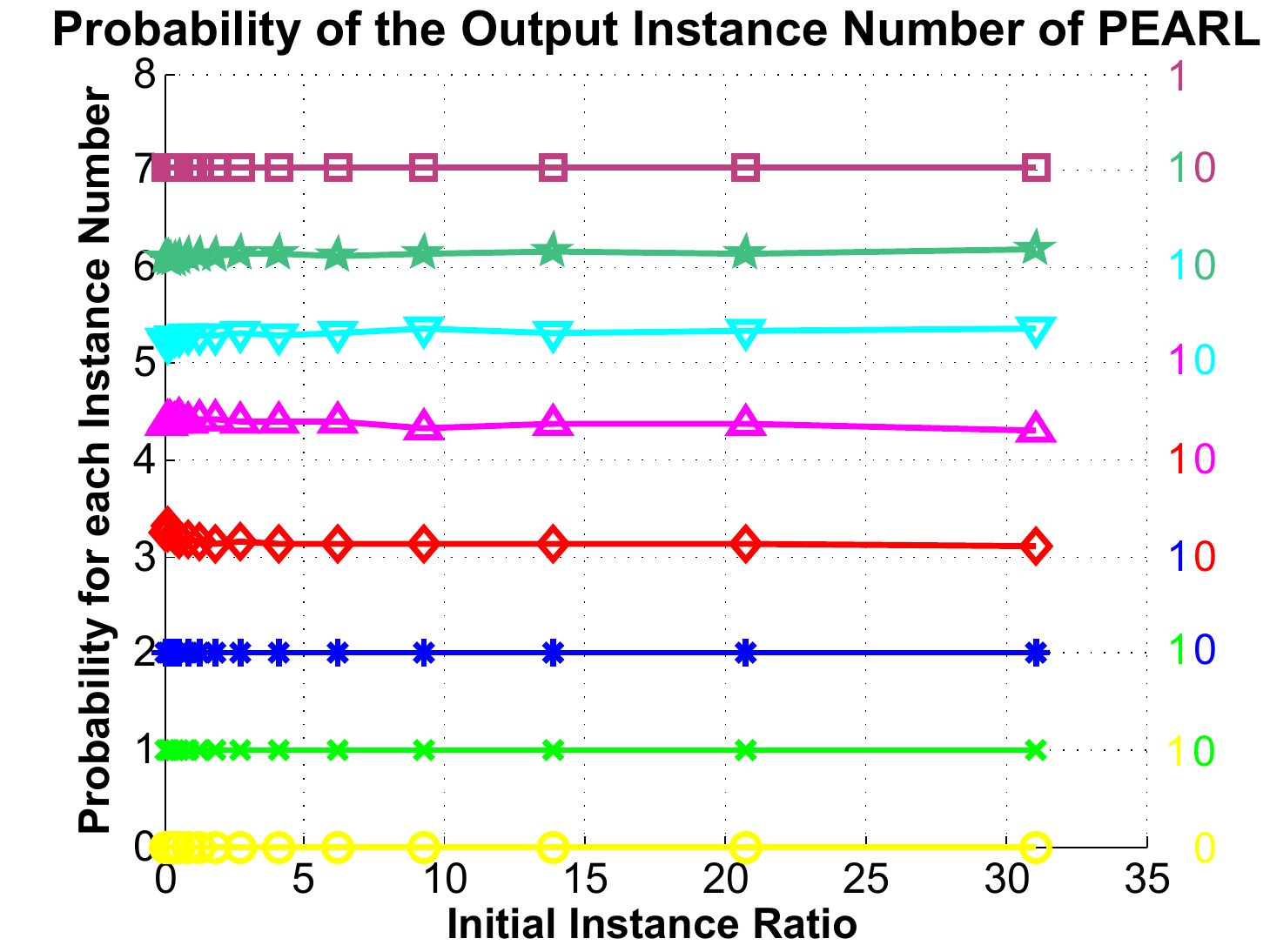}\includegraphics[width = 0.45\columnwidth]{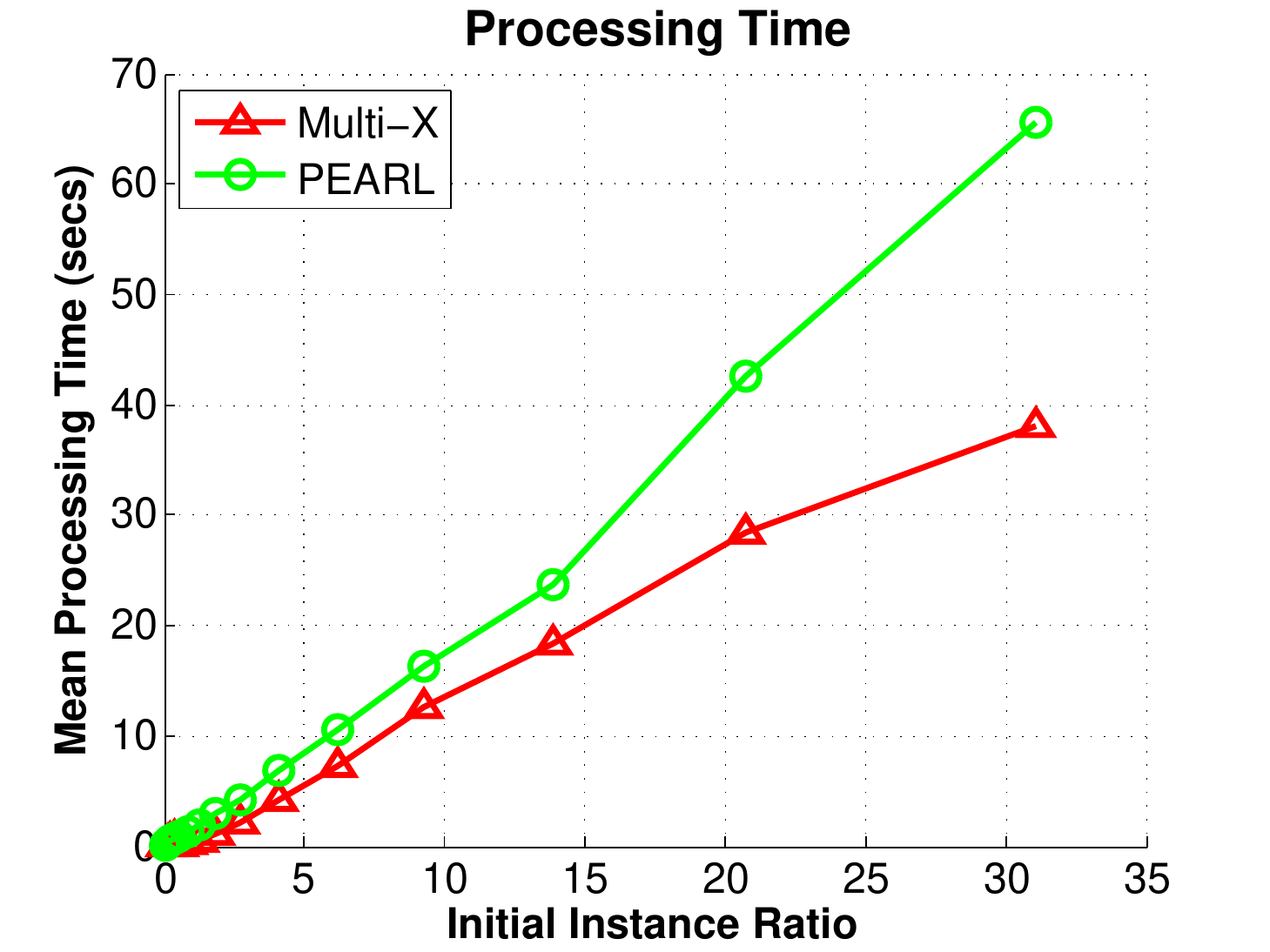}\\
    	\includegraphics[width = 0.45\columnwidth]{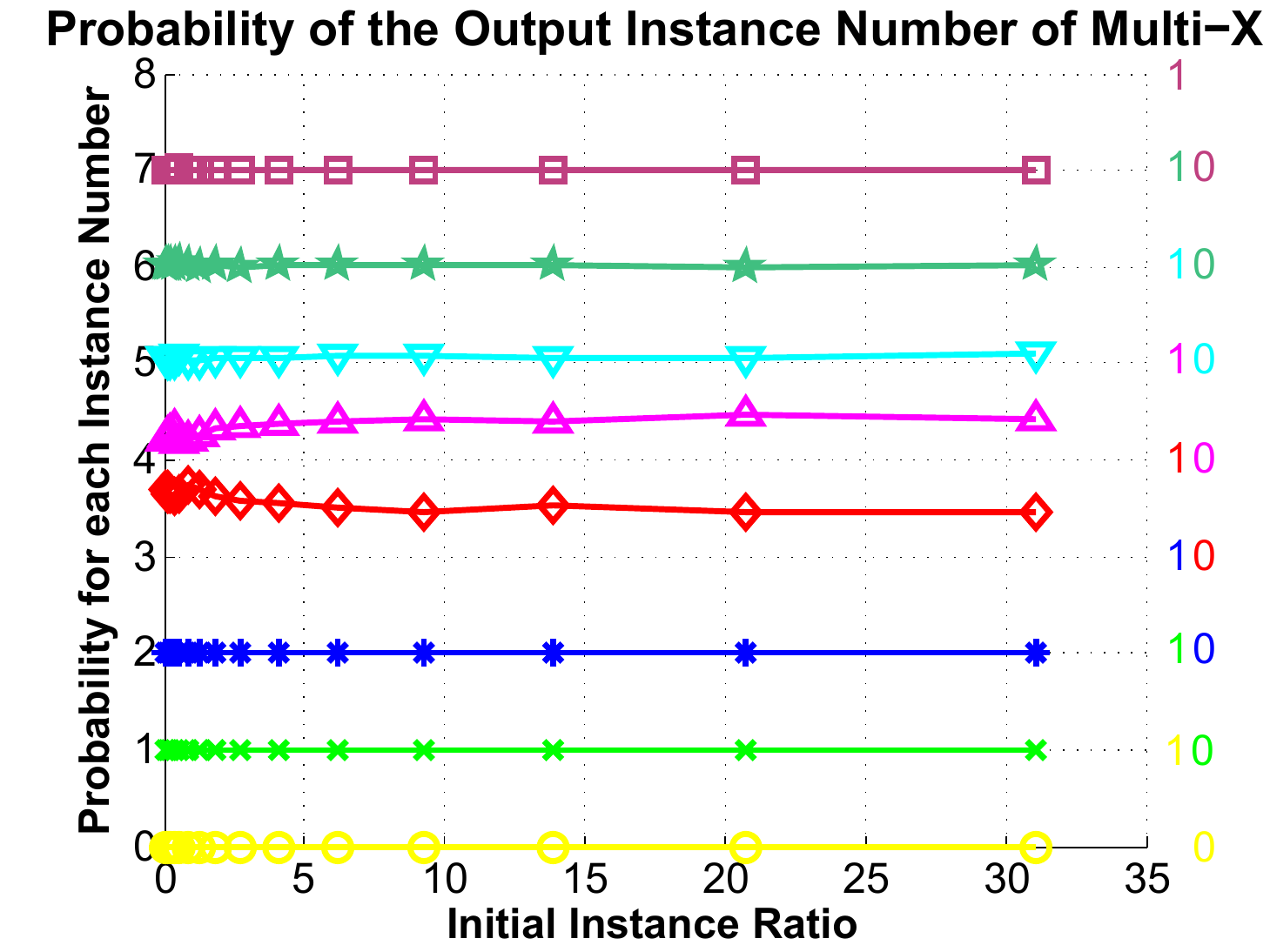}\includegraphics[width = 0.45\columnwidth]{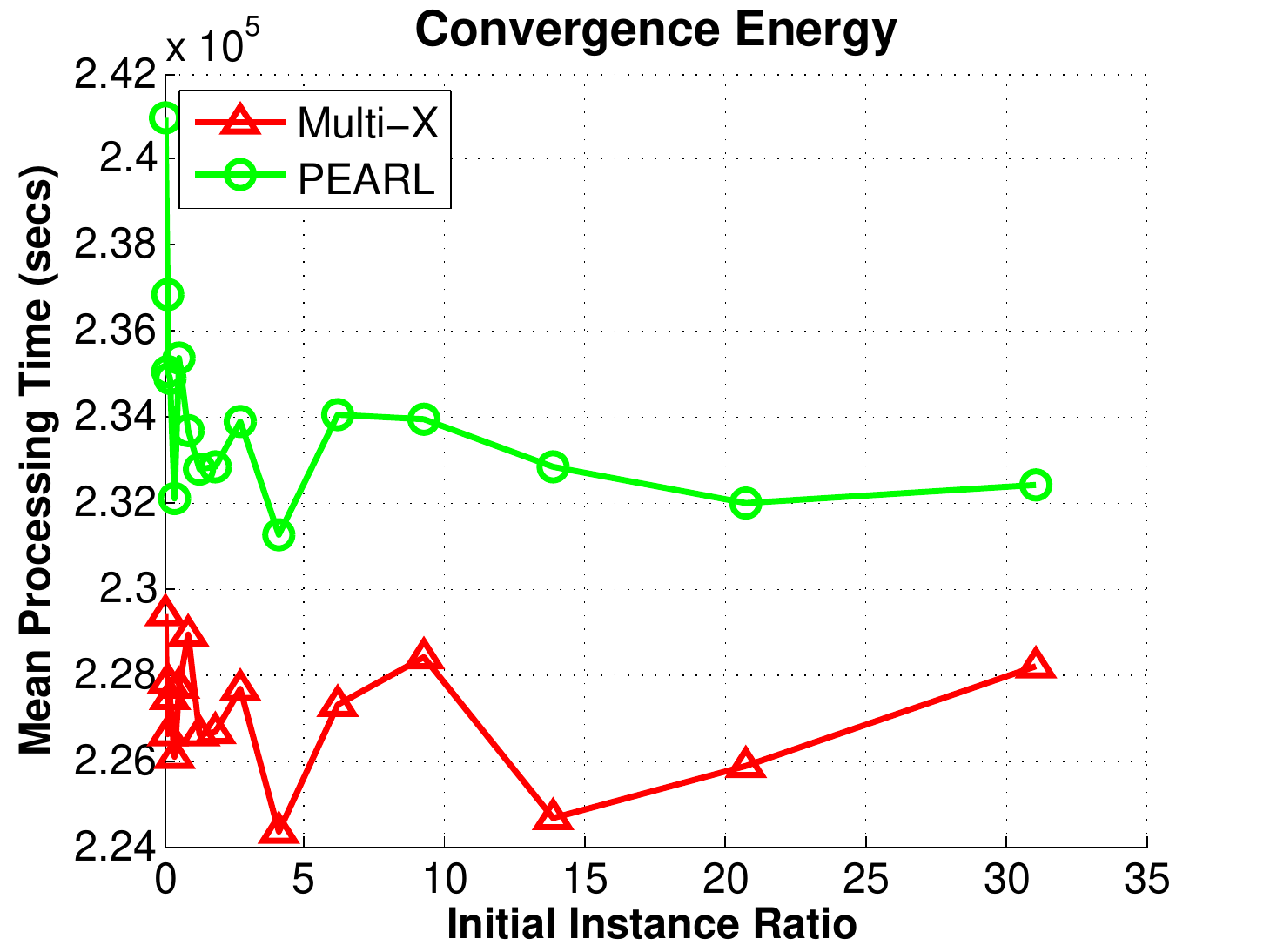}
        \caption{ \textit{Increasing instance number.}  Zero-mean Gaussian noise with $\sigma = 20$ pixels added to the point coordinates. (\textbf{Left}) the probability of returning 0, ..., 7 instances (vertical axis) for PEARL (top) and Multi-X (bottom) plotted as the function of the ratio of the initial instance number and the point number (horizonal axis). (\textbf{Right}): the processing time in seconds and convergence energy. }
      	\label{fig:pearl_vs_multix_a}
	\end{subfigure}
	\hfill
	\begin{subfigure}[t]{0.485\textwidth}
		\centering	
    	\includegraphics[width = 0.45\columnwidth]{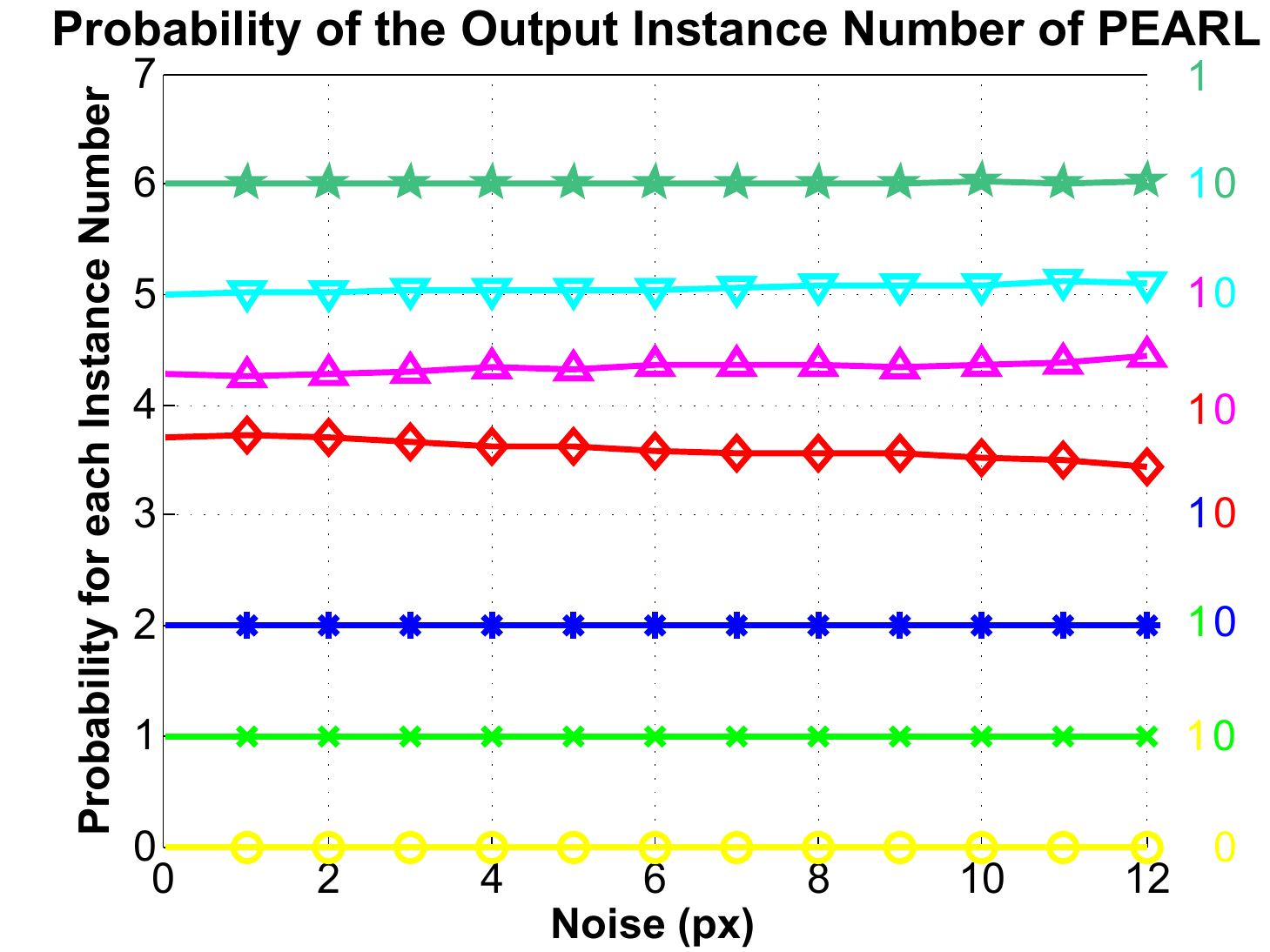}\includegraphics[width = 0.45\columnwidth]{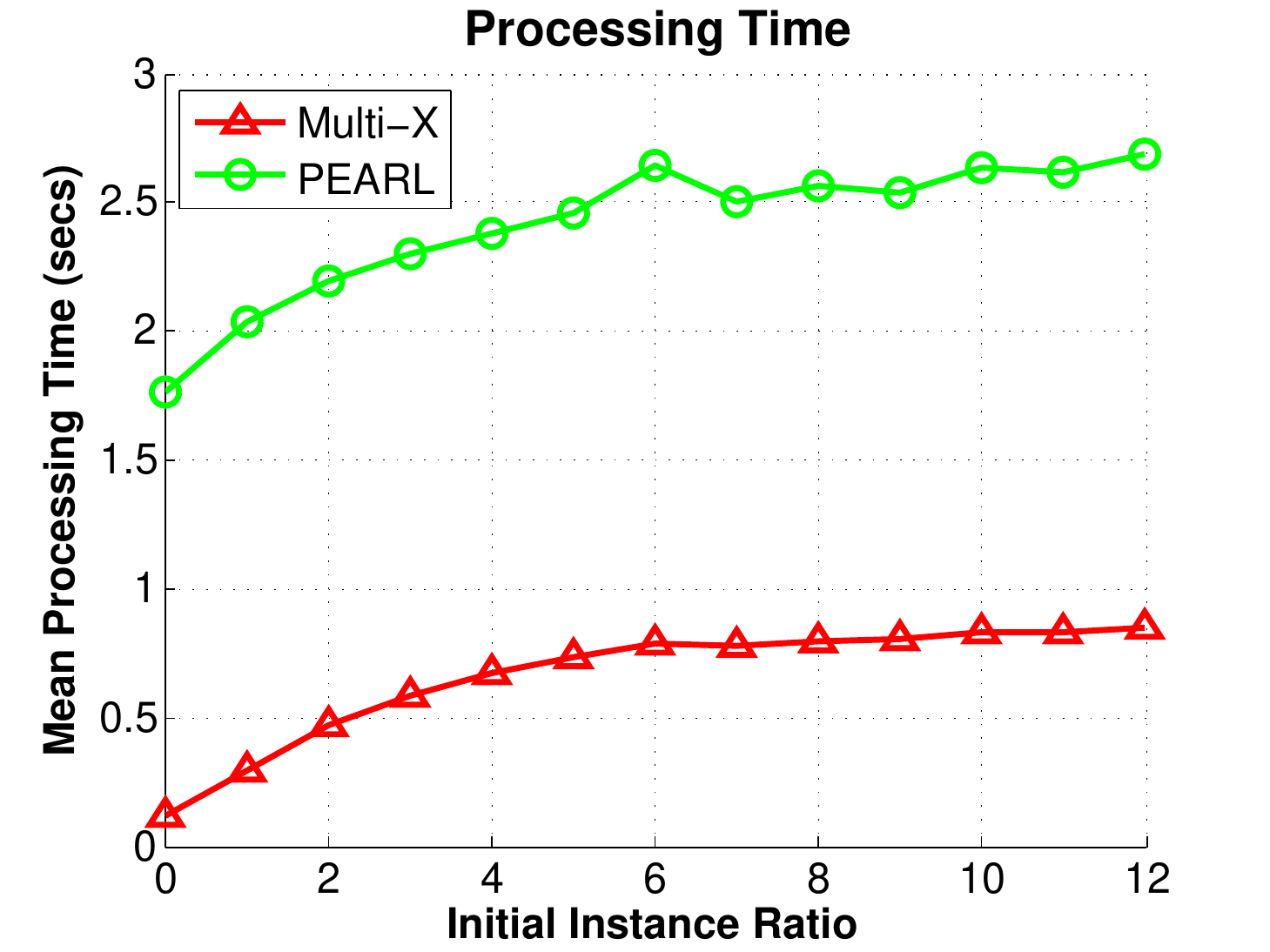}\\
    	\includegraphics[width = 0.45\columnwidth]{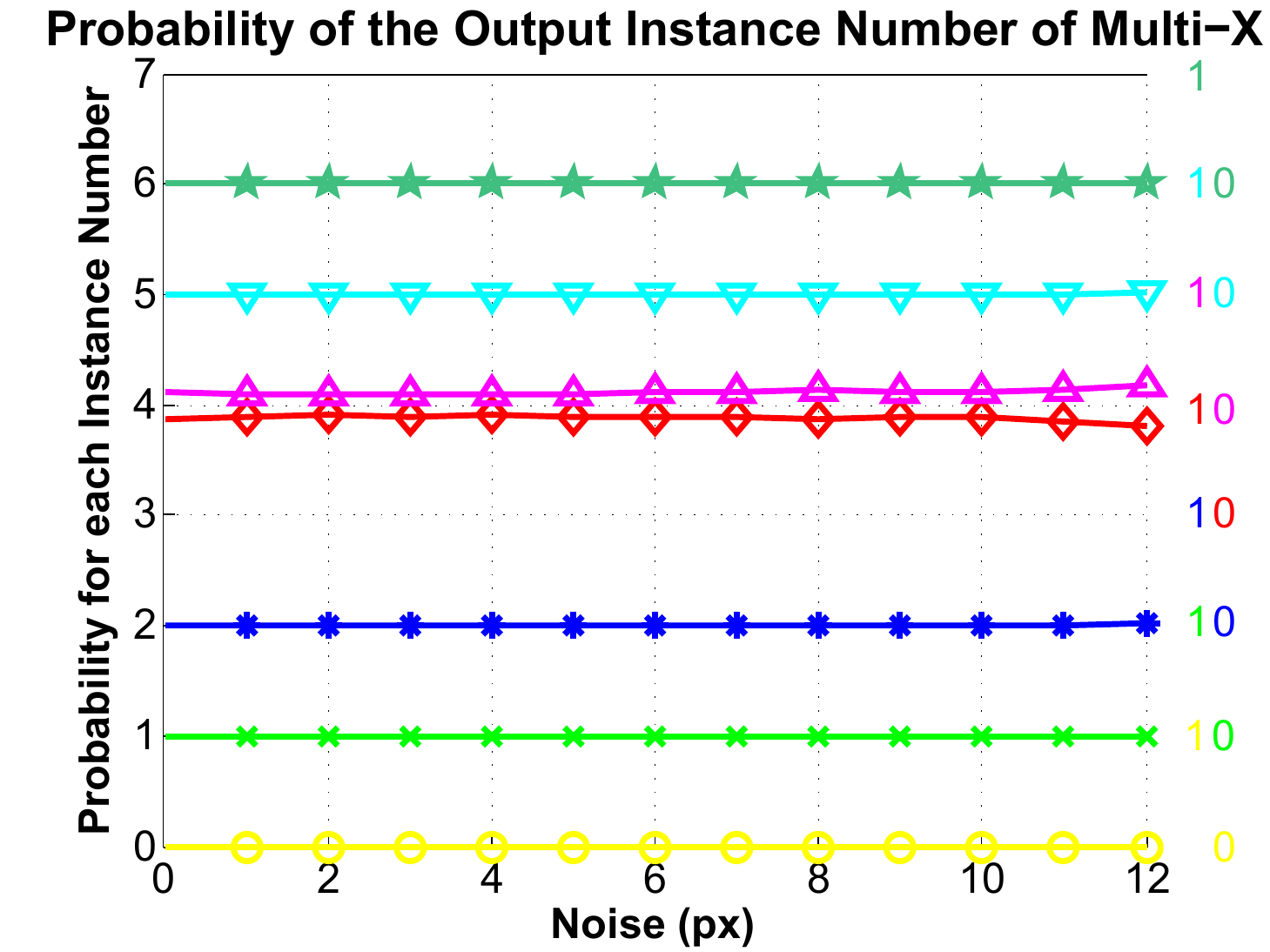}\includegraphics[width = 0.45\columnwidth]{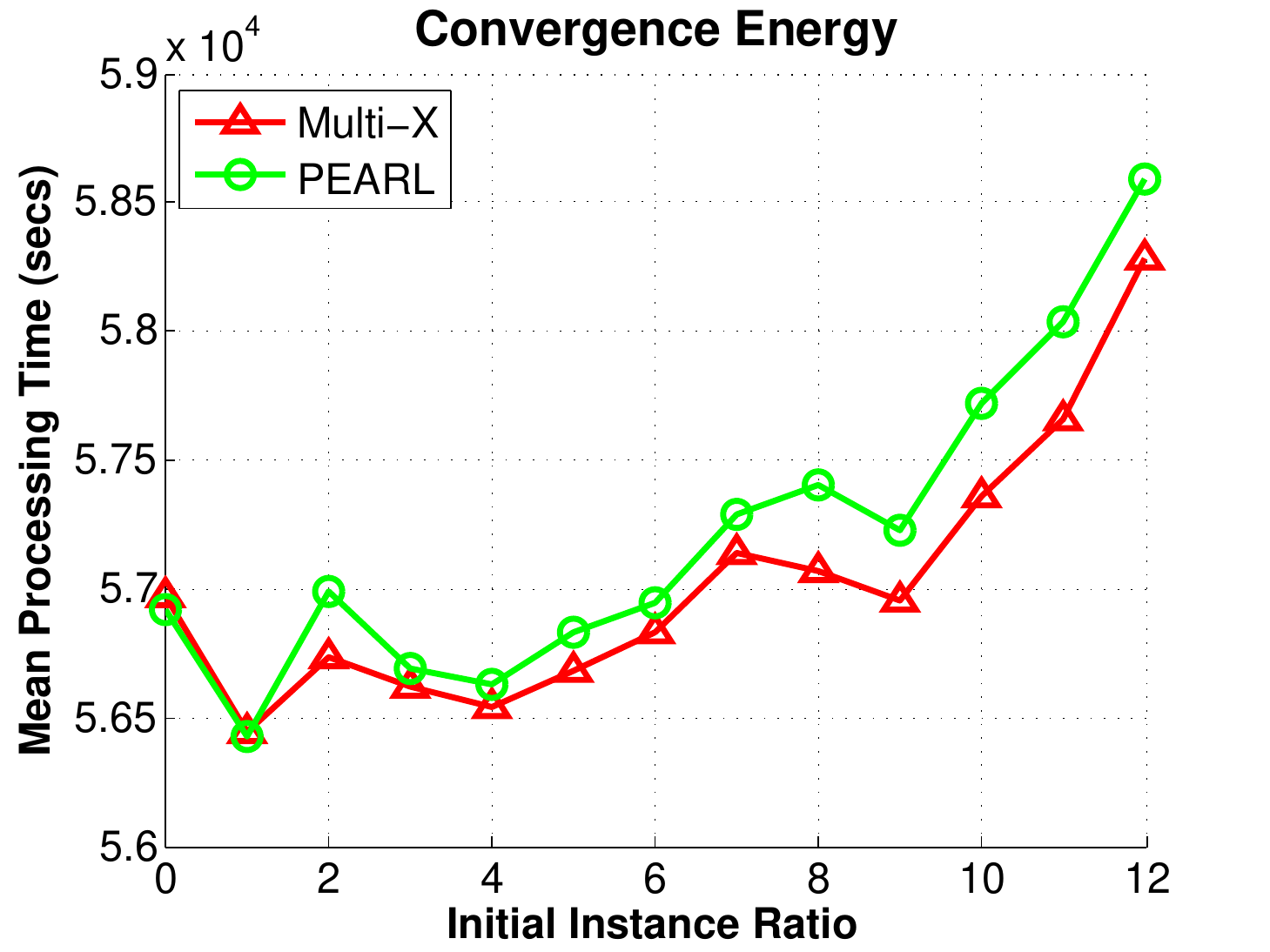}
        \caption{ \textit{Increasing noise.} The number of initial instances generated is twice the point number. (\textbf{Left}): the probability of returning instance numbers 0, ..., 7 (vertical axis) for PEARL (top) and Multi-X (bottom) plotted as the function of the noise $\sigma$ (horizonal axis). (\textbf{Right}): the processing time in seconds and convergence energy. }
      	\label{fig:pearl_vs_multix_b}
	\end{subfigure}
	\caption{ Comparison of PEARL and Multi-X. Three random lines sampled at $100$ locations, plus $200$ outliers. Parameters of both methods are: $h_{\text{max}} = 3$, and the outlier threshold is (a) 6 and (b) 3 pixels. }
	\label{fig:pearl_vs_multix}
\end{figure*}

Fig.~\ref{fig:pearl_vs_multix_a} shows the probability of returning an instance number for Multi-X (top-left) and PEARL (bottom-left). The numbers next to the vertical axis are the number of returned instances. The curve on their right shows the probability ($\in [0,1]$) of returning them. For instance, the red curve for PEARL on the right of number $3$ is close to the $0.1$ probability, while for Multi-X, it is approximately $0.6$. Therefore, Multi-X more likely returns the desired number of instances. The processing times (top-right), and convergence energies (bottom-right) are also reported. Values are plotted as the function of the initially generated instance number (horizontal axis; ratio w.r.t.\ to the input point number). 
The standard deviation of the zero-mean Gaussian-noise added to the point coordinates is $20$ pixels. Reflecting the fact that the noise $\sigma$ is usually not known in real applications, we set the outlier threshold to $6.0$ pixels. The maximum model number of the label cost was set to the ground truth value, $h_{\text{max}} = 3$, to demonstrate that suppressing instances exclusively with label cost penalties is not sufficient even with the proper parameters. It can be seen that Multi-X more likely returns the ground truth number of models, both its processing time and convergence energy are superior to that of PEARL.

For Fig.~\ref{fig:pearl_vs_multix_b}, the number of the generated instances was set to twice the point number, the threshold was set to $3$ pixels. Each reported property is plotted as the function of the noise $\sigma$ added to the point coordinates. The same trend can be seen as in Fig.~\ref{fig:pearl_vs_multix_a}: Multi-X is less sensitive to the noise than PEARL. It more often returns the desired instances, its processing time and convergence energy are lower. 

\paragraph{Simultaneous Line and Circle Fitting} is evaluated on 2D edges of banknotes and coins. Edges are detected by Canny edge detector and assigned to circles and lines manually to create a ground truth segmentation.\footnote{Submitted as supplementary material.}

Each method started with the same number of initial model instances: twice the data point (e.g.\ edge) number. The evaluated methods are PEARL~\cite{delong2012minimizing,isack2012energy}, T-Linkage~\cite{magri2014t}\footnote{ \url{http://www.diegm.uniud.it/fusiello/demo/jlk/}} and RPA~\cite{magri2015robust}\footnote{ \url{http://www.diegm.uniud.it/fusiello/demo/rpa/}} since they can be considered as the state-of-the-art and their implementations are available. PEARL and Multi-X fits circles and lines simultaneously, while T-Linkage and RPA sequentially. Table~\ref{tab:line_comparison} reports the number of false negative and false positive models. Multi-X achieved the lowest error for all test cases. 

\begin{table}
\center
  	\resizebox{\columnwidth}{!}{\begin{tabular}{l || c c | c c | c c  }
    \hline
    	& \multicolumn{2}{c}{(1)} & \multicolumn{2}{c}{(2)} & \multicolumn{2}{c}{(3)} \\ 
    \hline
 	 	 & FP & FN & FP & FN & FP & FN \\ 
    \hline
 	 	PEARL~\cite{isack2012energy} & 1 & \textbf{0} & 3 & \textbf{0} & 5 & 3 \\
 	 	T-Linkage~\cite{magri2014t} & \textbf{0} & 1 & 1 & 3 & \textbf{0} & 6 \\
 	 	RPA~\cite{magri2015robust} & \textbf{0} & 1 & \textbf{0} & 2 & \textbf{0} & 5 \\
 	 	\textbf{Multi-X} & \textbf{0} & \textbf{0} & \textbf{0} & \textbf{0} & \textbf{0} & \textbf{1} \\
	\hline
\end{tabular}}
\caption{ The number of false positive (FP) and false negative (FN) instances for simultaneous line and circle fitting.}
\label{tab:line_comparison}
\end{table}

\paragraph{Multiple Homography Fitting} is evaluated on the AdelaideRMF homography dataset~\cite{wongiccv2011} used in most recent publications. AdelaideRMF consists of $19$ image pairs of different resolutions with ground truth point correspondences assigned to planes (homographies). To generate initial model instances the technique proposed by Barath et al.~\cite{barath2016multi} is applied: a single homography is estimated for each correspondence using the point locations together with the related local affine transformations. Table~\ref{tab:homography_comparison} reports the results of PEARL~\cite{boykov2001fast}, FLOSS~\cite{lazic2009floss}, T-Linkage~\cite{magri2014t}, ARJMC~\cite{pham2011simultaneous}, RCMSA~\cite{pham2014random}, J-Linkage~\cite{toldo2008robust}, and Multi-X. To allow comparison with the state-of-the-art, all methods, including Multi-X, are tuned separately for each test and only the same $6$ image pairs are used as in~\cite{magri2014t}.

Results using a fixed parameter setting are reported in Table~\ref{tab:homography_table_fixed} (results, except that of Multi-X, copied from~\cite{magri2015robust}). Multi-X achieves the lowest errors. Compared to  results in Table~\ref{tab:homography_comparison} for parameters hand-tuned for each problem, the errors are significantly higher, but automatic parameter setting is the only possibility in many applications. Moreover, per-image-tuning leads to overfitting. 
\begin{figure}[htbp]
	\centering
	\includegraphics[width = 0.350\columnwidth]{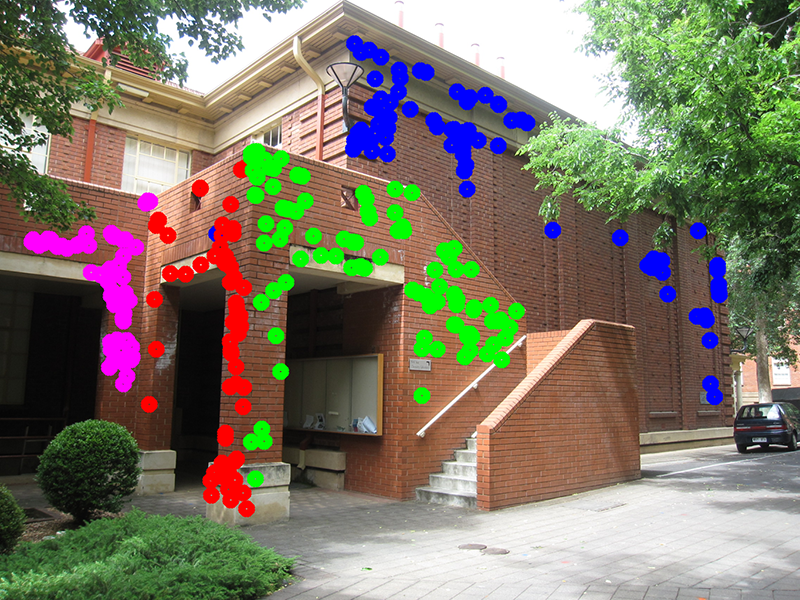}
    \includegraphics[width = 0.350\columnwidth]{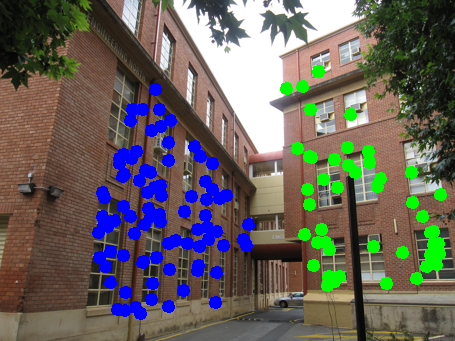}\\[1px]
    \includegraphics[width = 0.350\columnwidth]{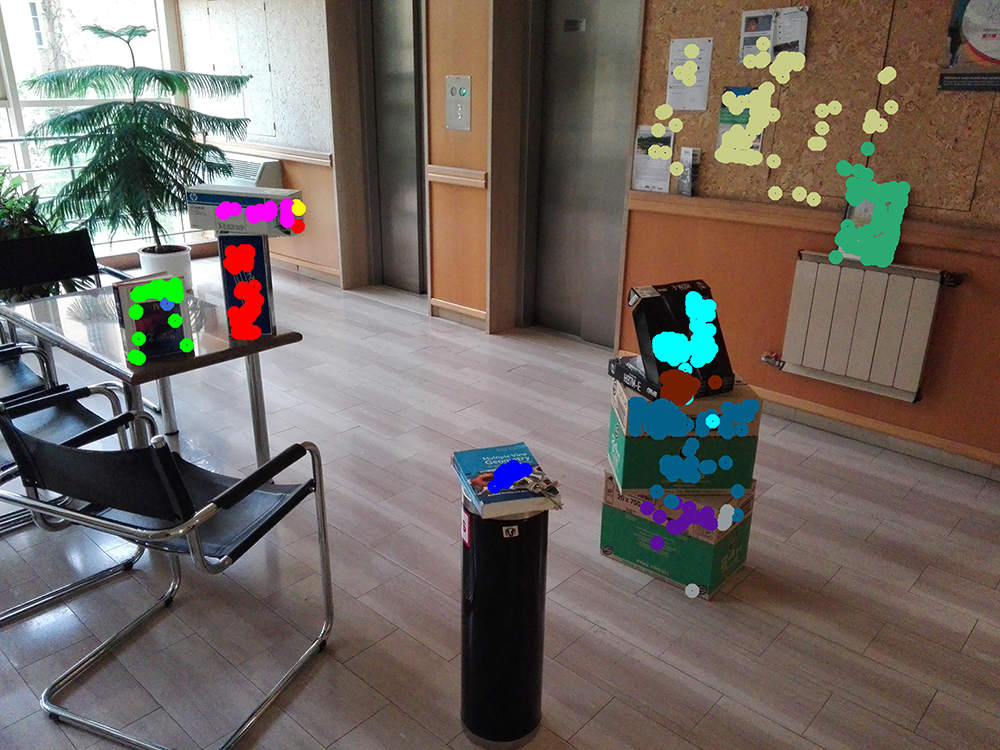} 
    \includegraphics[width = 0.350\columnwidth]{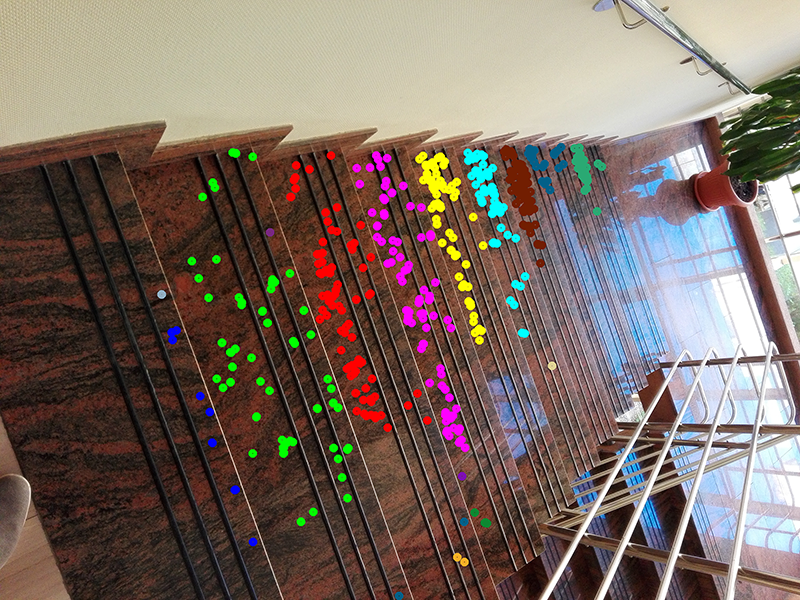} 
	\caption{ AdelaideRMF (top) and Multi-H (bot.) examples. Colors indicate the planes Multi-X assigned  points to.}
	\label{fig:motion_fig}
\end{figure}
\begin{table}
\center
\resizebox{\columnwidth}{!}{ \begin{tabular}{ c || c c c c c c c c }
 	\hline
 		& \rot{\# of planes} & 
        \rot{PEARL~\cite{isack2012energy}} & \rot{FLOSS~\cite{lazic2009floss}} & \rot{T-Lnkg~\cite{magri2014t}} & \rot{ARJMC~\cite{pham2011simultaneous}} & \rot{RCMSA~\cite{pham2014random}} & \rot{J-Lnkg~\cite{toldo2008robust}} & \rot{\textbf{Multi-X}} \\  	
 	\hline
 		(1) &4& 
        4.02 & 4.16 & 4.02 & 6.48 & 5.90 & 5.07 & \textbf{3.75} \\
 		(2) &6& 
        \hspace*{-1ex}18.18 &  \hspace*{-1ex}18.18 &  \hspace*{-1ex}18.17 &  \hspace*{-1ex}21.49 &  \hspace*{-1ex}17.95 &  \hspace*{-1ex}18.33 & \textbf{4.46} \\
 		(3) &2& 
        5.49 & 5.91 & 5.06 & 5.91 & 7.17 & 9.25 & \textbf{0.00} \\
 		(4) &3& 
        5.39 & 5.39 & 3.73 & 8.81 & 5.81 & 3.73 & \textbf{0.00} \\
 		(5) &2& 
        1.58 & 1.85 & 0.26 & 1.85 & 2.11 & 0.27 & \textbf{0.00} \\
 		(6) &2& 
        0.80 & 0.80 & 0.40 & 0.80 & 0.80 & 0.84 & \textbf{0.00} \\
	\hline
		{\footnotesize Avg.} & & 
        5.91 & 6.05 & 5.30 & 7.56 & 6.62 & 6.25 & \textbf{1.37} \\
		{\footnotesize Med.} & & 
        4.71 & 4.78 & 3.87 & 6.20 & 5.86 & 4.40 & \textbf{0.00} \\ 
	\hline
\end{tabular}}
\caption{ Misclassification error (\%) for the two-view plane segmentation on AdelaideRMF test pairs: (1) {\fontfamily{cmtt}\selectfont johnsonna}, (2) {\fontfamily{cmtt}\selectfont johnsonnb}, (3) {\fontfamily{cmtt}\selectfont ladysymon}, (4) {\fontfamily{cmtt}\selectfont neem}, (5) {\fontfamily{cmtt}\selectfont oldclassicswing}, (6) {\fontfamily{cmtt}\selectfont sene}. }
\label{tab:homography_comparison}
\end{table}

\begin{table}[h!]
	\centering
    \resizebox{\columnwidth}{!}{ \begin{tabular}{l || c c c c c }
    	\hline
    		& T-Lnkg & RCMSA & RPA & Multi-H & \textbf{Multi-X} \\
    		& \cite{magri2014t} & \cite{pham2014random} & \cite{magri2015robust} & \cite{barath2016multi} & \\
    	\hline
    		Avg. & 44.68 & 23.17 & 15.71 & 14.35 & \textbf{9.72} \\
	    	Med. & 44.49 & 24.53 & 15.89 & \phantom{x}9.56 & \textbf{2.49} \\  
    	\hline     
    \end{tabular} }	 
    \caption{ Misclassification errors (\%, average and median) for two-view plane segmentation on all the 19 pairs from AdelaideRMF test pairs using fixed parameters.}
    \label{tab:homography_table_fixed}
\end{table} 
%\vspace{0.2em}

\paragraph{Two-view Motion Segmentation} is evaluated on the AdelaideRMF motion dataset consisting of $21$ image pairs of different sizes and the ground truth -- correspondences assigned to their motion clusters.

Fig.~\ref{fig:motion_fig} presents example image pairs from the AdelaideRMF motion datasets partitioned by Multi-X. Different motion clusters are denoted by color. Table~\ref{tab:motion_table} shows comparison with state-of-the-art methods when all methods are tuned separately for each test case. Results are the average and minimum misclassification errors (in percentage) of ten runs. All results except that of Multi-X are copied from~\cite{wang2015mode}. For Table~\ref{tab:motion_table_fixed}, all methods use fixed parameters. For both test types, Multi-X achieved higher accuracy than the other methods. 
\begin{figure}[htbp]
	\centering
	\includegraphics[width = 0.350\columnwidth]{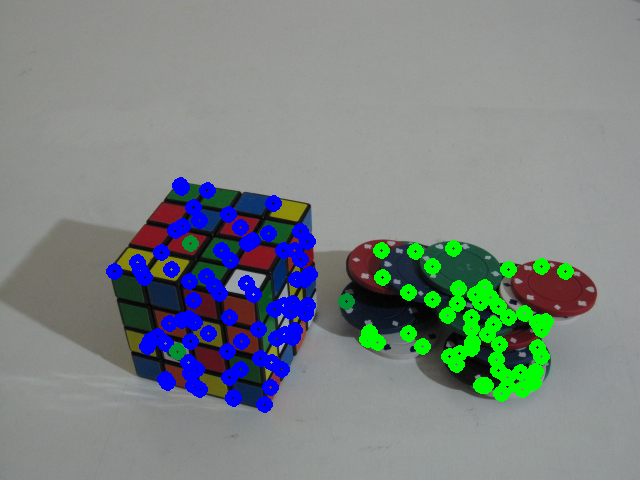}
    \includegraphics[width = 0.350\columnwidth]{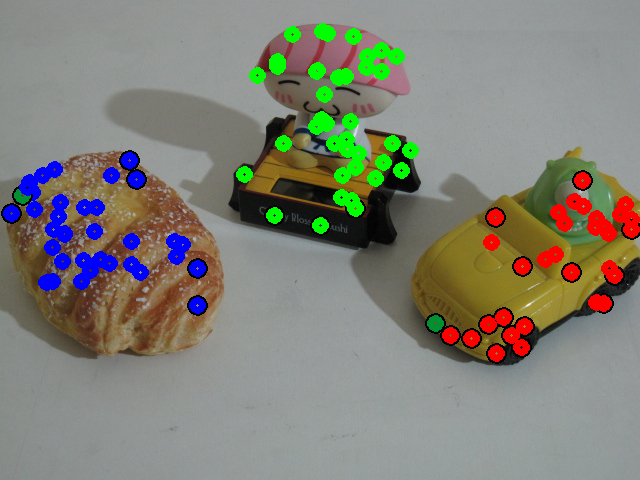}\\[1px]
    \includegraphics[width = 0.350\columnwidth]{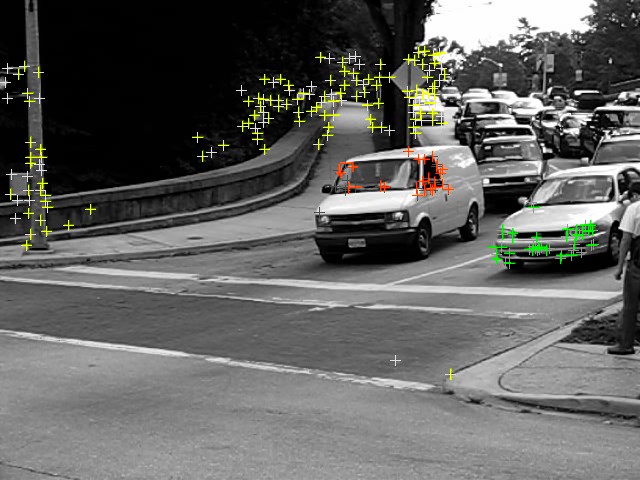} 
    \includegraphics[width = 0.350\columnwidth]{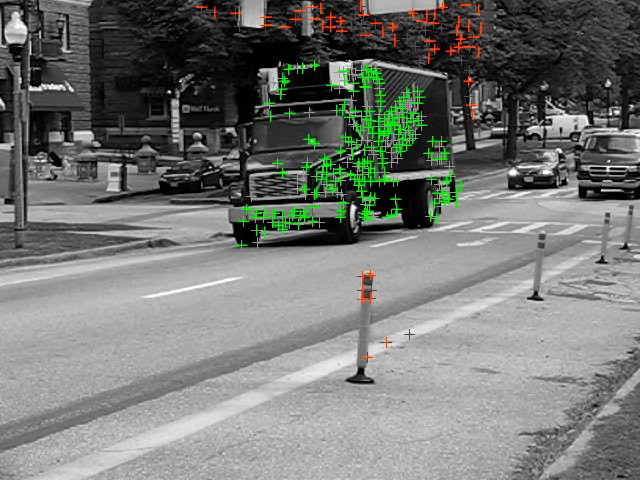} 
	\caption{ AdelaideRMF (top) and Hopkins (bot.) examples. Color indicates the motion Multi-X assigned a point to.}
	\label{fig:motion_fig}
\end{figure}
\begin{table*}[h!]
  \centering
  \begin{tabular}{l || c c | c c | c c | c c | c c | c c }
    \hline
    & \multicolumn{2}{c}{KF~\cite{chin2009robust}} & \multicolumn{2}{c}{RCG~\cite{liu2012efficient}} & \multicolumn{2}{c}{T-Lnkg~\cite{magri2014t}} & \multicolumn{2}{c}{AKSWH~\cite{tardif2009non}} & \multicolumn{2}{c}{MSH~\cite{wang2015mode}} & \multicolumn{2}{c}{\textbf{Multi-X}} \\
	& Avg. & Min. & Avg. & Min. & Avg. & Min. & Avg. & Min. & Avg. & Min. & Avg. & Min. \\
    \hline
	(1) & \phantom{x}8.42 & \phantom{x}4.23 & 13.43 & \phantom{x}9.52	 & 5.63 & 2.46 & \phantom{x}4.72 & \phantom{x}2.11 & 3.80 & 2.11 & \textbf{3.45} & \textbf{1.41} \\
	(2) & 12.53 & \phantom{x}2.81 & 13.35 & 10.92 & 5.62 & 4.82 & \phantom{x}7.23 & \phantom{x}4.02 & 3.21 & 1.61 & \textbf{2.27} & \textbf{0.40} \\
	(3) & 14.83 & \phantom{x}4.13 & 12.60 & \phantom{x}8.07 & 4.96 & 1.32 & \phantom{x}5.45 &\phantom{x} 1.42 & 2.69 & 0.83 & \textbf{1.45} & \textbf{0.41} \\
	(4) & 13.78 & \phantom{x}5.10 & \phantom{x}9.94 & \phantom{x}3.96 & 7.32 & 3.54 & \phantom{x}7.01 & \phantom{x}5.18 & 3.72 & 1.22 & \textbf{0.61} & \textbf{0.30} \\
	(5) & 16.87 & 14.55 & 26.51 & 19.54 & \textbf{4.42} & 4.00 & \phantom{x}9.04 & \phantom{x}8.43 & 6.63 & 4.55 & 5.24 & \textbf{1.80} \\
	(6) & 16.06 & 14.29 & 16.87 & 14.36 & 1.93 & 1.16 & \phantom{x}8.54 & \phantom{x}4.99 & 1.54 & 1.16 & \textbf{0.62} & \textbf{0.00} \\
	(7) & 33.43 & 21.30 & 26.39 & 20.43 & \textbf{1.06} & 0.86 & \phantom{x}7.39 & \phantom{x}3.41 & 1.74 & 0.43 & 5.32 & \textbf{0.00} \\
	(8) & 31.07 & 22.94 & 37.95 & 20.80 & 3.11 & 3.00 & 14.95 & 13.15 & 4.28 & 3.57 & \textbf{2.63} & \textbf{1.52} \\
    \hline
  \end{tabular}
  \caption{Misclassification errors (\%) for two-view motion segmentation on the AdelaideRMF dataset. All the methods were tuned separately for each video by the authors. Tested image pairs: (1) {\fontfamily{cmtt}\selectfont cubechips}, (2) {\fontfamily{cmtt}\selectfont cubetoy}, (3) {\fontfamily{cmtt}\selectfont breadcube}, (4) {\fontfamily{cmtt}\selectfont gamebiscuit}, (5) {\fontfamily{cmtt}\selectfont breadtoycar}, (6) {\fontfamily{cmtt}\selectfont biscuitbookbox}, (7) {\fontfamily{cmtt}\selectfont breadcubechips}, (8) {\fontfamily{cmtt}\selectfont cubebreadtoychips}. }
  \label{tab:motion_table}
\end{table*}
\begin{table}[h!]
	\centering
   \resizebox{\columnwidth}{!}{\begin{tabular}{l || c c c c c }
    	\hline
    		& RPA & RCMSA & T-Lnkg & AKSWH & \textbf{Multi-X} \\
    		& \cite{magri2015robust} & \cite{pham2014random} & \cite{magri2014t} & \cite{tardif2009non} & \\
    	\hline
    		Avg. & 5.62 & 9.71 & 43.83 & 12.59 & \textbf{2.97} \\
	    	Med. & 4.58 & 8.48 & 39.42 & 11.57 & \textbf{0.00} \\   
    	\hline     
    \end{tabular}}	
    \caption{Misclassification errors (\%, average and median) for two-view motion segmentation on all the 21 pairs from the AdelaideRMF dataset using fixed parameters.}
    \label{tab:motion_table_fixed}
\end{table}
%
%\vspace{0.2em}

\begin{figure}[htbp]
    \centering
    %\begin{subfigure}[b]{0.80\columnwidth}
        	\includegraphics[width = 0.70\columnwidth]{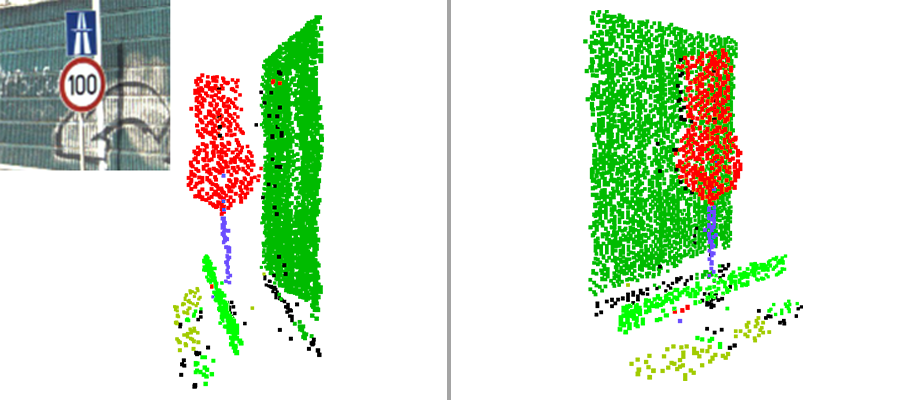}
    %\end{subfigure}
    %\begin{subfigure}[b]{0.80\columnwidth}
        	\includegraphics[width = 0.70\columnwidth]{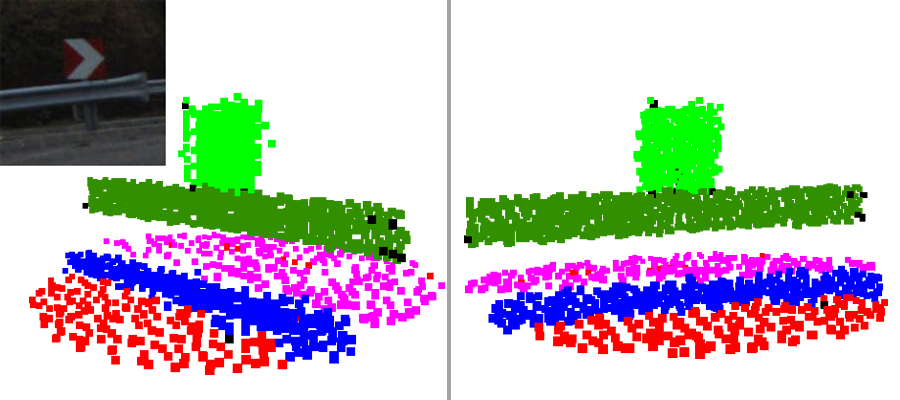}
    %\end{subfigure}    	
    \caption{ Results of simultaneous plane and cylinder fitting to LIDAR point cloud in two scenes. Segmented scenes visualized from different viewpoints. There is only one cylinder on the two scenes: the pole of the traffic sign on the top. Color indicates the instance Multi-X assigned a point to.}
    \label{fig:plane_fig}
\end{figure}

\paragraph{Simultaneous Plane and Cylinder Fitting} is evaluated on LIDAR point cloud data (see Fig.~\ref{fig:plane_fig}). The annotated database\footnote{It will be made available after publication.} consists of traffic signs, balusters and the neighboring point clouds truncated by a 3-meter-radius cylinder parallel to the vertical axis. Points were manually assigned to signs (planes) and balusters (cylinders). 

Multi-X is compared with the same methods as in the line and circle fitting section. PEARL and Multi-X fit cylinders and planes simultaneously while T-Linkage and RPA sequentially. Table~\ref{tab:plane_comparison} reports that Multi-X is the most accurate in all test cases except one. 

\begin{table}
\center
  	\resizebox{\columnwidth}{!}{\begin{tabular}{l || c c c c  }
	\hline
    	& PEARL~\cite{isack2012energy} & T-Lnkg~\cite{magri2014t} & RPA~\cite{magri2015robust} & \textbf{Multi-X} \\
    \hline
 		(1) & 10.63 & 57.46 & 46.83  & \hphantom{x}\textbf{8.89} \\
 		(2) & 10.88 & 41.79 & 53.39 & \hphantom{x}\textbf{4.72} \\
 		(3) & 37.34 & 52.97 & 61.64 & \hphantom{x}\textbf{2.84} \\
 		(4) & 38.13 & 38.91 & 41.41 & \textbf{19.38} \\
 		(5) & 17.20 & 51.83 & 53.34 & \textbf{16.83} \\
 		(6) & \textbf{17.35} & 61.77 & 51.21 & 21.72 \\
 		(7) & \hphantom{x}6.12 & 12.49 & 80.45 & \hphantom{x}\textbf{5.72} \\
	\hline
\end{tabular}}
\caption{ Misclassification error (\%) of simultaneous plane and cylinder fitting to LIDAR data. See Fig.~\ref{fig:plane_fig} for examples. }
\label{tab:plane_comparison}
\end{table}
%\vspace{0.2em}

\paragraph{Video Motion Segmentation} is evaluated on $51$ videos of the Hopkins dataset~\cite{tron2007benchmark}. Motion segmentation in video sequences is the retrieval of sets of points undergoing rigid motions contained in a dynamic scene captured by a moving camera. It can be considered as a subspace segmentation under the assumption of affine cameras. For affine cameras, all feature trajectories associated with a single moving object lie in a 4D linear subspace in $\mathbb{R}^{2F}$, where F is the number of frames~\cite{tron2007benchmark}.

\begin{table}
\center
  	 \resizebox{1.0\columnwidth}{!}{\begin{tabular}{l || l | c c c c c }
    \hline
    	& & (1) & (2) & (3) & (4) & (5) \\
    \hline
 		\multirow{2}{*}{SSC~\cite{elhamifar2009sparse}} & Avg. & 0.06 & 0.76 & 3.95 & 2.13 & 1.08 \\
      	& Med. & \textbf{0.00} & \textbf{0.00} & \textbf{0.00} & 2.13 & \textbf{0.00} \\
	\hline
 		\multirow{2}{*}{T-Lnkg~\cite{magri2014t}} & Avg. & 1.31 & 0.48 & 6.47 & 5.32 & 2.47 \\
      	& Med. & \textbf{0.00} & 0.19 & 2.38 & 5.32 & \textbf{0.00} \\
	\hline
 		\multirow{2}{*}{RPA~\cite{magri2015robust}} & Avg. & 0.14 & 0.19 & 4.41 & 9.11 & 1.42 \\
      	& Med. & \textbf{0.00} & \textbf{0.00} & 2.44 & 9.11 & \textbf{0.00} \\
	\hline
 		\multirow{2}{*}{Grdy-RC~\cite{magri2016multiple}} & Avg. & 7.48 & 28.65 & 8.75 & 14.89 & 10.91 \\
      	& Med. & \textbf{0.00} & 1.53 & 0.20 & 14.89 & \textbf{0.00} \\
	\hline
 		\multirow{2}{*}{ILP-RC~\cite{magri2016multiple}} & Avg. & 0.54 & 0.35 & 2.40 & 2.13 & 0.98 \\
      	& Med. & \textbf{0.00} & 0.19 & 1.30 & 2.13 & \textbf{0.00} \\
	\hline
 		\multirow{2}{*}{J-Lnkg~\cite{toldo2008robust}} & Avg. & 1.75 & 1.58  & 5.32 & 6.91 & 2.70 \\
      	& Med. & \textbf{0.00} & 0.34 & 1.30 & 6.91 & \textbf{0.00} \\
	\hline
 		\multirow{2}{*}{\textbf{Multi-X}} & Avg. & \textbf{0.05}& \textbf{0.09} & \textbf{0.32} & \textbf{1.06} & \textbf{0.16} \\
      	& Med. & \textbf{0.00} & \textbf{0.00} & \textbf{0.00} & \textbf{1.06} & \textbf{0.00} \\
	\hline
\end{tabular}}
\caption{ Misclassification errors (\%, average and median) for multi-motion detection on $51$ videos of Hopkins dataset: (1) {\fontfamily{cmtt}\selectfont Traffic2} -- 2 motions, $31$ videos, (2) {\fontfamily{cmtt}\selectfont Traffic3} -- $3$ motions, $7$ videos, (3) {\fontfamily{cmtt}\selectfont Others2} -- $2$ motions, $11$ videos, (4) {\fontfamily{cmtt}\selectfont Others3} -- $3$ motions, $2$ videos, (5) {\fontfamily{cmtt}\selectfont All} -- $51$ videos. }
\label{table:video_motion}
\end{table}

Table~\ref{table:video_motion} shows that the proposed method outperforms the state-of-the-art: SSC~\cite{elhamifar2009sparse}, T-Linkage~\cite{magri2014t}, RPA~\cite{magri2015robust}, Grdy-RansaCov~\cite{magri2016multiple}, ILP-RansaCov~\cite{magri2016multiple}, and J-Linkage~\cite{toldo2008robust}. Results, except for Multi-X, are copied from~\cite{magri2016multiple}. Fig.~\ref{fig:motion_fig} shows two frames of the tested videos. 

%\subsection{Indicator of Robustness}

%In our experience, the spatial coherence term $E_g$ is a good indicator of the model quality. In $68\%$ of the cases, selecting the solution with the lowest $E_g$ value out of three runs lead to the most accurate choice. In $78\%$, there was at most a $3\%$ difference in misclassification error between the best and the selected solution. As Multi-X is fast, it is feasible to apply it repeatedly%
%\footnote{Multi-X is a randomized algorithm - a property it inherits from the initialization stage.}
% and select the solution with the lowest $E_g$ value. 

\subsection{Processing Time}

 Multi-X is orders of magnitude faster than currently available Matlab implementations of J-Linkage, T-Linkage and RPA. Attacking the fitting problem with a technique similar to PEARL and SA-RCM, it is significantly faster since it benefits from high reduction of the number of instances in the Median-Shift step (see Table~\ref{table:times}).  

\begin{table}
\center
  	\resizebox{0.99\columnwidth}{!}{\begin{tabular}{r || c c | c c | c c | c c | c c }
    \hline
    	 & \multicolumn{2}{c}{(1)} & \multicolumn{2}{c}{(2)} & \multicolumn{2}{c}{(3)} & \multicolumn{2}{c}{(4)} & \multicolumn{2}{c}{(5)} \\
        \# &\textbf{M} & T & \textbf{M} & T & \textbf{M} & T & \textbf{M} & T & \textbf{M} & T \\
    \hline
        {\footnotesize 100} & 0.1 & \phantom{xx}0.4 & 0.1 & 0.3 & 0.1 & 0.3 & 0.0 & 0.2 & 0.1 & \phantom{xx}0.4 \\
        {\footnotesize 500} & 2.0 & \phantom{x}14.0 & 3.2 & 8.4 & 2.1 & 8.4 & 0.8 & 7.0 & 3.8 & \phantom{x}15.9 \\
        {\footnotesize 1000} & 5.1 & 102.8 & - & - & - & - & - & - & 7.5 & 120.9 \\
    \hline
\end{tabular}}
\caption{ Processing times (sec) of Multi-X (M) and T-Linkage (T) for the problem of fitting (1) lines and circles, (2) homographies, (3) two-view motions, (4) video motions, and (5) planes and cylinders. The number of data points is shown in the first column. }
\label{table:times}
\end{table}

\section{Conclusion}

%A novel formulation is proposed for multi-class multi-instance model fitting. It approaches the problem as an optimization of an energy originating from multiple sources. Multi-X extends PEARL algorithm~\cite{isack2012energy}, inter alia, with a new move in the label space: replacing a set of labels, i.e.\ instances, with the density mode in the model parameter domain. 
%
%The method outperforms the state-of-the-art in multiple homography, rigid motion, simultaneous plane and cylinder fitting; motion segmentation; and 2D edge interpretation (circle and line fitting). Multi-X runs in time approximately linear in the number of data points, it is an order of magnitude faster than available implementations of commonly used methods. \textit{The source code and the datasets for line-circle and plane-cylinder fitting will be made available with the publication.}

A novel multi-class multi-instance model fitting method has been proposed. It extends an energy minimization approach with a new move in the label space: replacing a set of labels corresponding to model instances by the mode of the density in the model parameter domain. Most of its key parameters are set adaptively making it applicable as a black box on a range of problems. 
Multi-X outperforms the state-of-the-art in multiple homography, rigid motion, simultaneous plane and cylinder fitting; motion segmentation; and 2D edge interpretation (circle and line fitting). Multi-X runs in time approximately linear in the number of data points, it is an order of magnitude faster than available implementations of commonly used methods. \textit{The source code and the datasets for line-circle and plane-cylinder fitting will be made available with the publication.}

{\small
\bibliographystyle{ieee}
\bibliography{egbib}
}

\end{document}